\documentclass[sigconf]{acmart}

\AtBeginDocument{%
  \providecommand\BibTeX{{%
    \normalfont B\kern-0.5em{\scshape i\kern-0.25em b}\kern-0.8em\TeX}}}

\copyrightyear{2023}
\acmYear{2023}
\setcopyright{acmlicensed}\acmConference[KDD '23]{Proceedings of the 29th ACM SIGKDD Conference on Knowledge Discovery and Data Mining}{August 6--10, 2023}{Long Beach, CA, USA}
\acmBooktitle{Proceedings of the 29th ACM SIGKDD Conference on Knowledge Discovery and Data Mining (KDD '23), August 6--10, 2023, Long Beach, CA, USA}
\acmPrice{15.00}
\acmDOI{10.1145/3580305.3599953}
\acmISBN{979-8-4007-0103-0/23/08}

\usepackage{subfigure}
\usepackage{multirow}
\usepackage{bm}
\usepackage{bbm}
\usepackage{dsfont}
\usepackage{color,soul}
\usepackage{listings}
\usepackage{xcolor}

\usepackage[linesnumbered,vlined,ruled,noend]{algorithm2e}
\usepackage{xspace}
\newcommand{\sys}{Rover\xspace}

\definecolor{codegray}{rgb}{0.5,0.5,0.5}

\usepackage{enumitem}

\usepackage{algorithmic}

\begin{document}

\title{\sys: An Online Spark SQL Tuning Service via Generalized Transfer Learning}

\author{Yu Shen$^{*}$}
\affiliation{%
  \institution{School of CS, Peking University}
  \institution{ByteDance Inc.}
  \city{Beijing}
  \country{China}
}
\email{shenyu@pku.edu.cn}

\author{Xinyuyang Ren$^{*}$}
\affiliation{%
  \institution{ByteDance Inc.}
  \city{Beijing}
  \country{China}
}
\email{renxinyuyang@bytedance.com}

\author{Yupeng Lu}
\affiliation{%
  \institution{School of CS, Peking University}
  \institution{ByteDance Inc.}
  \city{Beijing}
  \country{China}
}
\email{xinkelyp@pku.edu.cn}

\author{Huaijun Jiang}
\affiliation{%
  \institution{Center for Data Science, Peking University}
  \institution{ByteDance Inc.}
  \city{Beijing}
  \country{China}
}
\email{jianghuaijun@pku.edu.cn}

\author{Huanyong Xu}
\affiliation{%
  \institution{ByteDance Inc.}
  \city{Beijing}
  \country{China}
}
\email{xuhuanyong@bytedance.com}

\author{Di Peng}
\affiliation{%
  \institution{ByteDance Inc.}
  \city{Beijing}
  \country{China}
}
\email{pengdi@bytedance.com}

\author{Yang Li}
\affiliation{%
  \institution{School of CS, Peking University}
  \city{Beijing}
  \country{China}
}
\email{liyang.cs@pku.edu.cn}

\author{Wentao Zhang}
\affiliation{%
  \institution{Mila - Québec AI Institute}
  \city{Montréal}
  \country{Canada}
}
\email{wentao.zhang@mila.quebec}

\author{Bin Cui}
\affiliation{%
  \institution{School of CS, Peking University}
  \institution{Institute of Computational Social Science, Peking University (Qingdao)}
  \city{Beijing}
  \country{China}}
\email{bin.cui@pku.edu.cn}




\renewcommand{\shortauthors}{Shen et al.}
\renewcommand{\authors}{Yu Shen, Xinyuyang Ren, Yupeng Lu, Huaijun Jiang, Huanyong Xu, Di Peng, Yang Li, Wentao Zhang, Bin Cui}
\renewcommand{\thefootnote}{}


\begin{abstract}

Distributed data analytic engines like Spark are common choices to process massive data in industry.
However, the performance of Spark SQL highly depends on the choice of configurations, where the optimal ones vary with the executed workloads.
Among various alternatives for Spark SQL tuning, Bayesian optimization (BO) is a popular framework that finds near-optimal configurations given sufficient budget, but it suffers from the re-optimization issue and is not practical in real production.
When applying transfer learning to accelerate the tuning process, we notice two domain-specific challenges: 1) most previous work focus on transferring tuning history, while expert knowledge from Spark engineers is of great potential to improve the tuning performance but is not well studied so far; 2) history tasks should be carefully utilized, where using dissimilar ones lead to a deteriorated performance in production.

In this paper, we present \sys, a deployed online Spark SQL tuning service for efficient and safe search on industrial workloads.
To address the challenges, we propose generalized transfer learning to boost the tuning performance based on external knowledge, including expert-assisted Bayesian optimization and controlled history transfer. 
Experiments on public benchmarks and real-world tasks show the superiority of \sys over competitive baselines.
Notably, \sys saves an average of 50.1\% of the memory cost on 12k real-world Spark SQL tasks in 20 iterations, among which 76.2\% of the tasks achieve a significant memory reduction of over 60\%.
\let\thefootnote\relax\footnote{* Equal contribution.}

\end{abstract}





\settopmatter{printfolios=true}

\maketitle

\section{Introduction}
\label{sec:intro}
Big data query systems, like Hive~\cite{thusoo2009hive},  Pig~\cite{gates2009building}, Presto~\cite{sethi2019presto}, and Spark SQL~\cite{armbrust2015spark}, have been extensively applied in industry to efficiently process massive data for downstream business, such as recommendation and advertisement.
As a module of Apache Spark~\cite{zaharia2016apache}, Spark SQL inherits the benefits of Spark programming model~\cite{zaharia2010spark} and provides powerful integration with the Spark ecosystem.

However, it's often challenging to set proper configurations for optimal performance when executing Spark SQL tasks~\cite{herodotou2020survey}.
For example, the parameter \texttt{spark.executor.memory} specifies the amount of memory for an executor process.
A too large value leads to long garbage collection time while a too small value may cause out-of-memory errors.
Therefore, it's crucial to tune the configurations to achieve satisfactory performance (in terms of memory, runtime, etc.).
In this paper, we focus on tuning online Spark SQL tasks.
Unlike offline tuning where it is tolerable to simulate various configurations in a non-production cluster~\cite{bei2015rfhoc,herodotou2011starfish,lama2012aroma,yu2018datasize}, each configuration can only be evaluated in real production due to high overhead of offline evaluations or security concerns.
Thus, the configuration should be carefully selected to achieve high performance quickly (efficient) and low risks of negative results (safe).

Recent studies~\cite{alipourfard2017cherrypick,xin2022locat,fekry2020tuneful} apply the Bayesian optimization (BO) framework~\cite{hutter2011sequential,bergstra2011algorithms,li2021openbox,shen2022divbo} to reduce the required number of evaluations to find a near-optimal configuration.
In brief, BO trains a surrogate on evaluated configurations and their performance, and then selects the next configuration by balancing exploration and exploitation.
However, BO suffers from the re-optimization issue~\cite{li2022transbo}, i.e., BO always starts from scratch when tuning a new task.
For periodic tasks that only execute once a day, it may take months until BO finds a good configuration, which makes it impractical to deploy in industry.
To accelerate this tuning process, the TLBO (transfer learning for BO) community studies to integrate external knowledge into the search algorithm.
But for online Spark SQL tuning, we notice some domain-specific challenges and describe them as two questions below,

\textbf{C1. What to transfer?} Among literature of the TLBO community, almost all methods~\cite{feurer2018scalable,li2022transbo,li2022transfer,golovin2017google,perrone2018scalable} consider utilizing the tuning history of previous tasks, which is suitable for most scenarios. 
While for Spark SQL scenarios, we notice that domain-specific knowledge can also be utilized to further improve the tuning performance, e.g., parameter adjustment suggestions from experienced engineers.
However, how to translate the knowledge to code and how to combine them with the automatic search algorithm are still open questions.

\textbf{C2. How to transfer?} Most previous work~\cite{feurer2018scalable,li2022transbo,swersky2013multi,yogatama2014efficient,wistuba2015sequential} assume that similar tasks always exist in history and consider only one or all tasks (often less than 20).
However, in real production, history tasks accumulate over time, and most of them may be dissimilar to the current task.
Utilizing dissimilar tasks may lead to deteriorated tuning performance (i.e., negative transfer), which is especially intolerable in online tuning.
Therefore, it's also essential to identify similar history tasks beforehand and design a proper solution to combine the knowledge from various similar tasks.

In this paper, we propose \sys, a deployed online Spark SQL tuning service that aims at efficient and safe search on in-production workloads.
To address the re-optimization issue with the aforementioned challenges, we propose generalized transfer learning to boost the tuning performance based on external knowledge.
For the ``what to transfer'' question, we first apply expert knowledge to design a compact search space and then integrate expert knowledge into the automatic search algorithm to boost the performance in both the initialization and search phases.
For the ``how to transfer'' question, we design a regression model for history filtering to avoid negative results and build a strong ensemble to enhance the final performance.
Our contributions are summarized as follows,

1. We introduce \sys, an online Spark SQL tuning service that provides user-friendly interfaces and performs an efficient and safe search on given workloads.
So far, \sys has already been deployed in ByteDance big data development platform and is responsible for tuning 10k+ online Spark SQL tasks simultaneously.


2. To address the two challenges in algorithm design, we integrate expert knowledge into the automatic search algorithm  (\textbf{C1}).
In addition, we enhance the final performance and avoid negative results by filtering dissimilar history tasks and building a strong ensemble (\textbf{C2}).

3. Extensive experiments on public benchmarks and real-world tasks show that \sys clearly outperforms state-of-the-art tuning frameworks used for Spark and databases.
Notably, compared with configurations given by Spark engineers, \sys saves an average of 50.1\% of the memory cost on 12k real-world Spark SQL tasks within 20 iterations, among which 76.2\% of the tasks achieve a significant memory reduction of over 60\%.

\section{Related Work}
\label{sec:related}
Spark SQL~\cite{armbrust2015spark} is a widely applied module in Apache Spark~\cite{zaharia2016apache} for high-performance structured data processing. 
Despite its comprehensive functionality, the performance of a Spark SQL application is controlled by more than 200 parameters.
The Spark official website~\cite{spark_doc}, as well as various industry vendors such as Cloudera~\cite{cloudera} and DZone~\cite{dzone}, provide heuristic instructions to tune those parameters, and previous work~\cite{wang2015performance,singhal2018performance,venkataraman2016ernest,zacheilas2017dione,sidhanta2019deadline,lan2021survey} also propose to optimize heuristically designed cost models instead of the real objectives.
However, these methods still requires users of a thorough understanding of the system mechanism.

To reduce the barrier of expert knowledge and further improve the heuristically tuned results, previous methods~\cite{bao2018learning,gounaris2017dynamic,petridis2016spark,yu2018datasize,zhu2017bestconfig,cheng2021tuning} propose to train a performance model based on a large number of evaluations using an offline cluster.
However, they incur a large computational overhead and can not adapt to potential data change in online scenarios.

Recently, a variety of methods propose advanced designs of the tuning framework, which consequently reduce the number of evaluations to find satisfactory configurations and can be tentatively applied to online tuning.
Some methods combine the performance model with the genetic algorithm to derive better configurations. 
RFHOC~\cite{bei2015rfhoc} uses several random forests to model each task and applies the genetic algorithm to suggest configurations. DAC~\cite{yu2018datasize} combines hierarchical regression tree models with the genetic algorithm and applies the datasize-aware design to capture workload change.
Other methods leverage the Bayesian Optimization (BO) framework~\cite{snoek2012practical,hutter2011sequential,bergstra2011algorithms} and achieve state-of-the-art performance in Spark tuning. CherryPick~\cite{alipourfard2017cherrypick} directly performs BO on a discretized search space. 
LOCAT~\cite{xin2022locat} further combines dynamic sensitivity analysis and datasize-aware Gaussian process (GP) to perform optimization on important parameters.
Despite the competitive converged results, the aforementioned methods suffer from the re-optimization issue~\cite{li2022transbo}, which is, the performance model needs retraining and still requires a number of online configuration evaluations for each coming task.
To alleviate the issue, Tuneful~\cite{fekry2020tuneful,peters2019tune} is a pioneering work that attempts to apply a multi-task GP~\cite{swersky2013multi} to utilize the most similar previous task in Spark tuning.
However, even considering literature from the TLBO (transfer learning for BO) community, most work~\cite{feurer2015initializing,feurer2018scalable,li2022transbo,li2022transfer,bai2023transfer} only consider history tasks as useful knowledge to transfer.
In addition, those methods assume that there always exists similar history tasks for new tasks. 
A mechanism to pre-select similar tasks and prevent negative transfer when similar tasks are absent is also required in real production.

In the database community, many ML-based methods~\cite{zhang2019end,li2019qtune,van2017automatic,zhang2021restune,kunjir2020black,ma2018query} have been proposed to search for the best database knobs.
While sharing a similar spirit, some of the methods can be directly applied in online Spark tuning.
OtterTune~\cite{van2017automatic} and ResTune~\cite{zhang2021restune} adopt the BO framework, where ResTune integrates the knowledge of all history tasks to further improve its performance.
Due to the difference in research fields, in this paper, we mainly focus on tuning parameters for Spark SQL tasks.

\section{Overview}
In this section, we introduce the overview of \sys, and then provide preliminaries of the algorithm design, which are problem definition, Bayesian optimization framework and expert rule trees.

\subsection{System Overview}

\begin{figure*}
  \begin{center}
	\includegraphics[width=1\linewidth]{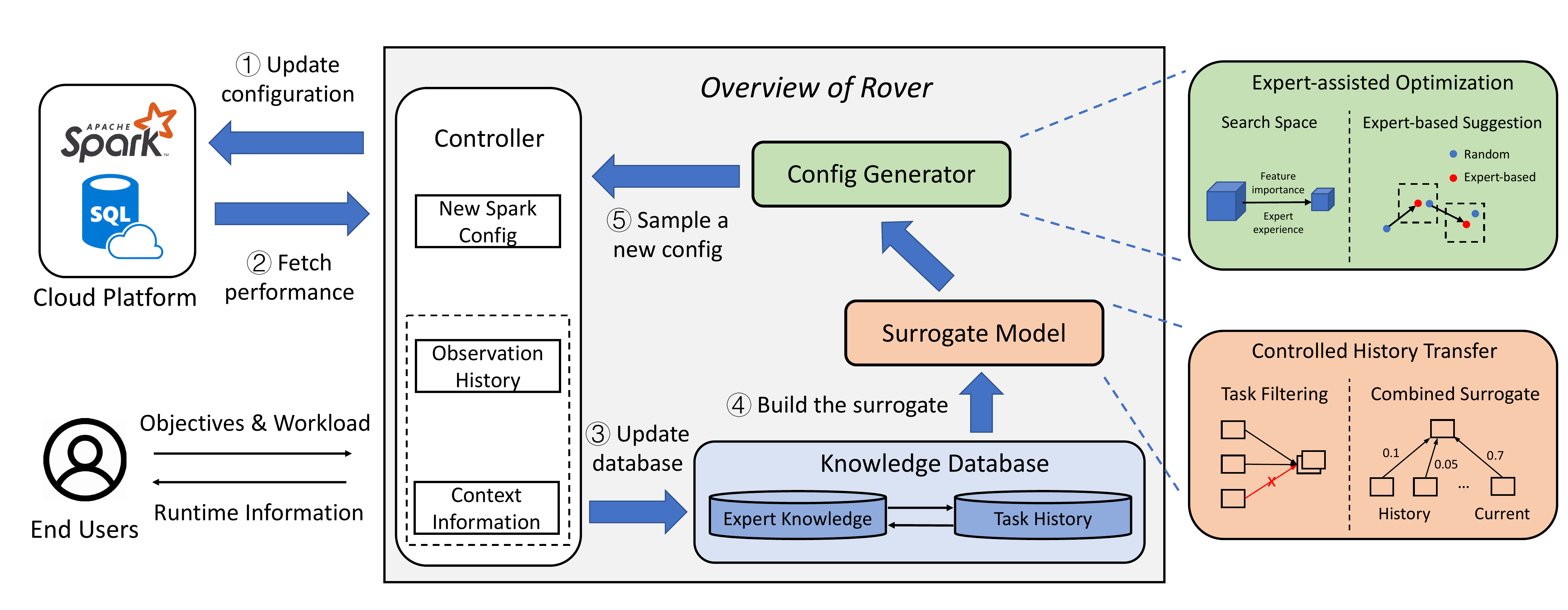}
  \end{center}
  \caption{\sys architecture and its interaction with Spark and users.}
 \label{fig:overview}
\end{figure*}

Figure~\ref{fig:overview} shows the overview of \sys. The framework contains the following four components:
1) the \sys controller interacts with end users and the cloud platform, and controls the tuning process; 
2) the knowledge database stores the configuration observations and context information on different Spark SQL tasks, and the expert knowledge concluded from previous tasks;
3) the surrogate model fits the relationship between configurations and observed performance;
4) the configuration generator suggests a promising configuration for the tuning task during each iteration.

To start a tuning task, the end user first specifies the tuning objective and uploads the workload to the controller.
Then the iterative workload starts in \sys. 
During each iteration, the \textbf{\sys controller} updates a new configuration to the cloud platform and fetches the specified objective and context information after the workload finishes.
The information is sent to the \textbf{knowledge database}, where a database stores all the observations and information of previous tuning tasks.
Based on expert knowledge and task history, a \textbf{surrogate model} is built to fit the relationship between configurations and objectives.
The \textbf{configuration generator} then samples a new configuration to the \sys controller based on the prediction of the surrogate model.
Note that, all configurations are run in the real production environment, which means \sys does not require additional budgets for offline tuning as in previous work~\cite{bao2018learning,petridis2016spark,yu2018datasize}.
When the tuning task ends, the cloud platform applies the best configuration found during optimization.
\sys also provides a set of user interfaces, where the end users can monitor the runtime status and make further decisions on whether to early stop a task or increase the tuning budget.

\subsection{Problem Definition}
The Spark configuration has a great impact on the performance of a Spark SQL task~\cite{alipourfard2017cherrypick,fekry2020tuneful}.
A configuration refers to a set of parameters with a certain choice of value.
Given a Spark SQL tuning task, the goal of \sys is to find a near-optimal Spark configuration that minimizes a pre-defined objective function.
The problem definition is as follows,
\begin{equation}
    x^*=\arg\min_{x \in \mathcal{X}} \mathcal{L}(x),
\label{eq:definition}
\end{equation}
where $\mathcal{X}$ is the Spark configuration search space and $\mathcal{L}$ is the optimization target. 
For targets that need to be maximized, we simply use its negative form in Equation~\ref{eq:definition}.
Unlike some previous methods~\cite{alipourfard2017cherrypick} where $\mathcal{L}$ is a fixed target, \sys supports multiple optimization targets, e.g., memory, CPU utilization, and complex targets like costs (runtime $\times$ prices per hours).

The Spark SQL tuning tasks are typically black-box optimization problems, which means the actual objective value $\mathcal{L}(x)$ of a configuration $x$ can not be obtained unless we run it in the real cluster.
A straightforward solution is to try a sufficient number of configurations and apply the best one.
However, in online industrial scenarios, it's impossible to evaluate a large number of Spark configurations due to the limited budget. 
To reduce the number of evaluated Spark configurations perform an efficient online search, \sys adopts the Bayesian optimization framework, which will be introduced in the following section.

\begin{algorithm}[t]
  \caption{Pseudo code for Bayesian optimization}
  \label{algo:bo}
  \begin{algorithmic}[1]
  \REQUIRE the search budget $\mathcal{B}$, the configuration search space $\mathcal{X}$, \\
  the initial configurations $X_{init}$.
  \ENSURE the best observed Spark configuration.
  \STATE evaluate the initial configurations and initialize observations $D=\{(x_{init},y_{init})|x_{init} \in X_{init}\}$.
  \WHILE{budget $\mathcal{B}$ does not exhaust} 
  \STATE build the surrogate $M$ based on observations $D$.
  \STATE select the next configuration $x_n$ that maximizes the acquisition function $a(x; M)$.
  \STATE evaluate the selected configuration and obtain its performance $y_n$.
  \STATE augment $D=D\cup(x_n,y_n)$.
  \ENDWHILE
  \STATE \textbf{return} the best observed configuration.

\end{algorithmic}
\end{algorithm}

\subsection{Bayesian optimization framework}
Bayesian optimization (BO) is a popular optimization framework designed to solve black-box problems with expensive evaluation costs.
The main advantage of BO is that it estimates the uncertainty of unseen configurations and balances exploration and exploitation when selecting the next configuration to evaluate.
BO follows the framework of sequential model-based optimization, which loops over the following three steps
as shown in Algorithm~\ref{algo:bo}
:
1) BO fits a probabilistic surrogate model $M$ based on the observations $D=\{(x_1, y_1),...,(x_{n-1}, y_{n-1})\}$, in which $x_i$ is the configuration evaluated in the $i^{th}$ iteration and $y_i$ is its corresponding observed performance; 2) BO uses the surrogate $M$ to select the most promising configuration $x_n$ by maximizing 
$x_{n}=\arg\max_{x \in \mathcal{X}}a(x; M)$, where $a(x; M)$ is the acquisition function designed to balance the trade-off between exploration and exploitation; 3) BO evaluates the configuration $x_n$ to obtain $y_n $(i.e., evaluate the 
Spark configuration in the real cluster and obtains the objective value), and augment the observations by $D=D\cup\{(x_n,y_n)\}$. 

\noindent\textbf{Surrogate.}
Following popular implementations in the BO community, \sys applies the Gaussian Process (GP) as the surrogate.
Compared with other alternatives, Gaussian Process is parameter-free and computes the posterior distributions in closed forms.
Given an unseen configuration $x$, GP outputs a posterior marginal Gaussian distribution, whose predictive mean and variance are formulated as follows,
\begin{equation}
\begin{aligned}
    \mu(x)&=K(X,x)(K(X, X)+\tau^2I)^{-1}Y,\\
    \sigma^2(x)&=K(x, x)+\tau^2I-(K(X, X)+\tau^2I)^{-1}K(X, x),
\end{aligned}
\end{equation}
where $K$ is the covariance matrix, $X$ are the observed configurations, $Y$ are the observed performance of $X$, and $\tau^2$ is the level of white noise. In practice, we use the Matern 5/2 kernel when computing covariance in Gaussian Process.

\noindent\textbf{Acquisition Function.}
The acquisition function determines the most promising configuration to evaluate in the next iteration. 
We use the Expected Improvement (EI) function, which measures the expected improvement of a configuration over the best-observed performance, which is formulated as,
\begin{equation}
    EI(\bm x)= \int_{-\infty}^{\infty} \max(y^*-y, 0)p_{M}(y|\bm x)dy,
\label{eq:ei}
\end{equation}
where $y^*$ is the best performance observed so far. 

\noindent\textbf{Context-aware Design.}
In online scenarios, the running environment (context) of a tuning task often changes due to data shifts or other processes running in the shared cluster. 
While the optimal configuration slightly differs according to the running environment, we follow the surrogate implementation in previous work~\cite{zhang2021restune}, which also takes the data size and context vector from SparkEventLog as inputs. 
Similar to the configuration vector, we also use the Matern 5/2 kernel for the context vector.
With this extension, the surrogate can deal with dynamic workloads in online scenarios.

\subsection{Expert Rule Trees}
\label{sec:rules}
Spark experts often adjust the values of specific Spark parameters when the observed performance meets a given condition.
To simulate this behavior and translate the knowledge to code, \sys defines a set of actions, which we refer to as expert rule trees. 
An expert rule includes the following five parts: 1) Parameter name, which determines the Spark parameter to adjust; 2) Direction, which shows whether to increase or decrease the value of the parameter if a change is needed; 3) Condition, which decides whether to adjust the parameter; 4) Step, which decides how much to change the parameter if a change is needed; 5) Bounds, which ensures that the parameter will not exceed limits for safety consideration.

We take a practical expert rule as an example. 
To run a Spark SQL task using the MapReduce programming structure, it's essential to decide the number of mappers to process the input data.
If the number is set too large, the execution of a mapper finishes too quickly, which may lead to high time overhead to start and delete the mappers, and high memory overhead to maintain those mappers.
To prevent creating too many mappers, experts can reduce the number of mappers by increasing the value of \texttt{maxPartitionBytes}.
An expert rule example can be: Double (Direction \& Step) its value (Parameter) if the average execution time of mappers is less than 12 seconds (Condition). The minimal and maximal bounds of the value are 16M and 4G, respectively (Bounds).
In \sys, we design 47 rules in total.
Please refer to the Appendix~\ref{appendix:rules} for more details.


\section{Algorithm Design}
\label{sec:algo}
As mentioned in Section~\ref{sec:intro}, \sys designs a generalized transfer learning framework for Spark SQL tuning tasks.
To answer the "what to transfer" question, we will first introduce expert-assisted Bayesian optimization to show how \sys integrates expert knowledge in each part of the Bayesian optimization algorithm. 
And then, we will answer the "how to transfer" question by introducing how we improve previous transfer learning methods from knowledge filtering and combination via controlled history transfer.

\subsection{Expert-assisted Bayesian Optimization}
\label{sec:expert-assisted}
While Bayesian optimization requires random initialization to accumulate sufficient observations to fit a surrogate model, the performance may be far from satisfactory at the beginning of the optimization in online tuning scenarios.
Meanwhile, previous methods have shown performance improvement in early iterations by using deliberately designed heuristic rules in Spark tuning scenarios\cite{spark_doc}.
However, expert knowledge often finds a moderate configuration quickly, but its performance is less promising than BO given a larger search budget. 
Then, it seems natural and intuitive to speed up Spark SQL tuning by combining the advantages of both Bayesian optimization and expert knowledge, i.e., achieving promising results with fewer evaluations. 
By analyzing the BO framework, \sys proposes to strengthen the BO framework with heuristic expert knowledge in the following three parts:

\subsubsection{Compact Search Space.}
\label{sec:search_space}
The size of the search space greatly influences the optimization, as adding a dimension in the search space leads to an exponential growth of possible Spark configurations. 
To reduce the difficulty of fitting a BO surrogate, \sys tunes a Spark SQL task using a heuristic search space with fixed parameters.
Concretely, we collect historical tasks that run on the entire search space, and then use SHAP~\cite{lundberg2020local}, a toolkit for measuring feature importance, to rank the importance of each Spark parameter over the observed performance.
The final search space used in online tuning includes 10 parameters, 
which are the top-5 important five parameters given by SHAP and 5 extra important parameters selected by Spark SQL experts.
The detailed parameters are provided in the Appendix~\ref{appendix:space}.

\begin{figure}
  \begin{center}
	\includegraphics[width=0.9\linewidth]{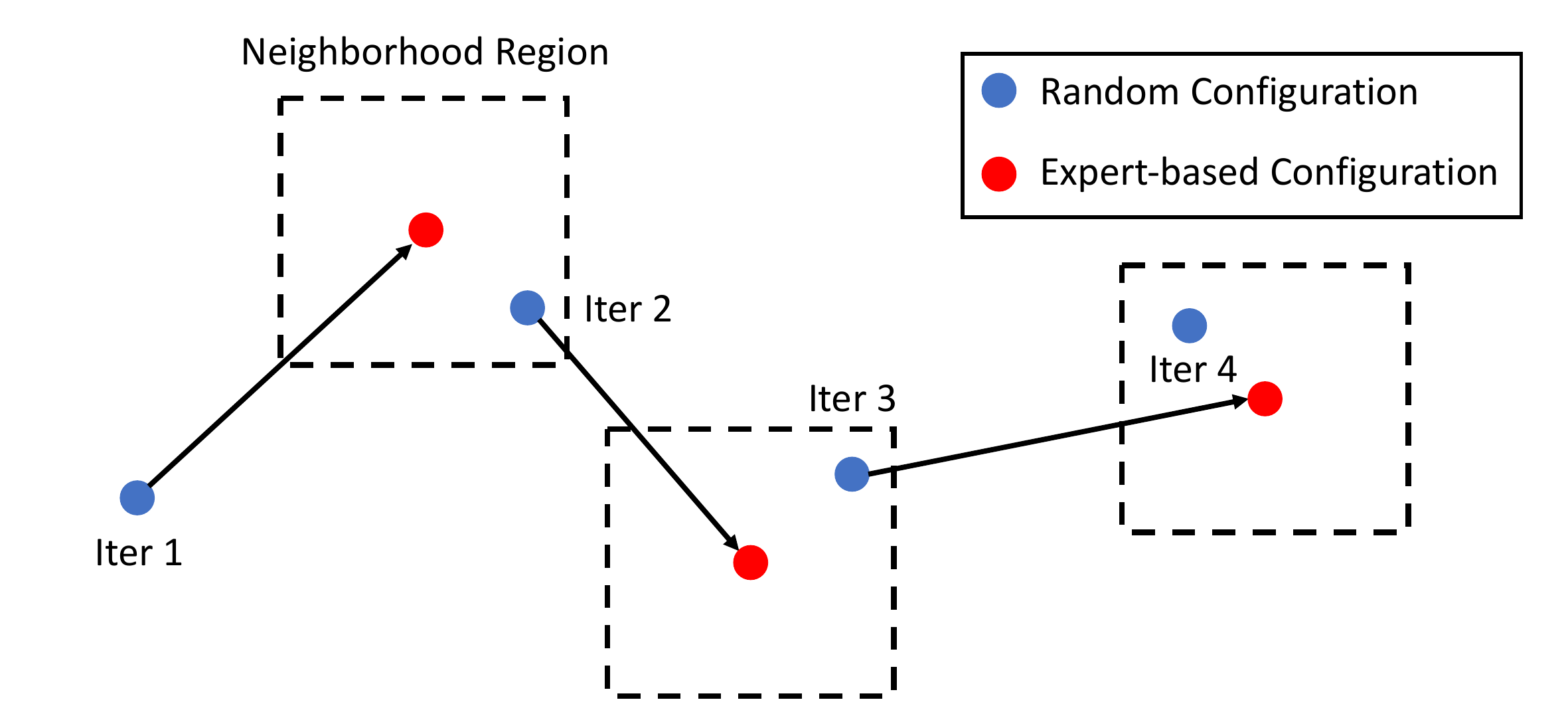}
  \end{center}
  \caption{An example of expert-based random initialization. Three configurations are randomly sampled from the neighborhood regions of expert-based configurations one by one.}
 \label{figure:initialization}
\end{figure}

\subsubsection{Expert-based Random Initialization.}
\label{sec:init}
While Bayesian optimization uses random start points for initialization, \sys improves this step by integrating the expert rule trees.
An example of the initialization phase is shown in Figure~\ref{figure:initialization}. 
During each initialization round, \sys first applies the expert rule trees on the previous configuration and obtains the updated one (the red point). 
Then, a new configuration (the blue point) is randomly chosen from the neighborhood region of the updated configuration.
In practice, the neighborhood region is set to be a region of configurations whose difference is less than $\pm 20\%$ in each parameter.

The initialization design combines both expert rules and random initialization, which has the following advantages: 
1) Compared with vanilla random initialization, expert rules tend to find configurations with better performance in early iterations; 
2) Compared with using expert rules alone, the random design ensures that each Spark parameter has different values among configurations. Since it is quite often that a Spark parameter dissatisfies all corresponding expert rules so that its value does not change, the random design at least changes the value of all parameters, which helps the BO surrogate to measure their importance easier;
3) Compared with random initialization, the neighborhood region prevents \sys from evaluating potentially bad configurations sampled in the entire search space, which ensures safe optimization in online scenarios.

\subsubsection{Expert-based Configuration Generator.}
\label{sec:config_gen}
Previous work reports that BO may suffer from over-exploitation in later iterations, i.e., the configurations are generated from a quite small region.
To address this issue, mature BO framework~\cite{jiang2021automated,feurer2015efficient} often adopts the $\epsilon$-greedy strategy, where there is a probability of $\epsilon$ that the configuration is suggested randomly instead of using the acquisition function.
Meanwhile, right after the initialization phase, the BO surrogate may not fit well based on scarce observations for initialization, which consequently leads to an unpromising configuration suggested by the BO acquisition function.
To address both two issues, we set a dynamic possibility that the expert rule trees generate the next configuration instead of the BO acquisition function. 
The intuition is that, if the BO surrogate performs worse, the expert rule trees will be applied more frequently.

Concretely, \sys assigns two weights to the expert rule trees and BO surrogate to control the probability of applying each component.
For expert rule trees, the weight is decreasing and is only controlled by the number of iterations.
The weight is formulated as $w_e=0.5^T+0.2$, where $T$ is the current number of iterations.
This matches the intuition that 1) the expert rule trees should be applied less frequently when the BO surrogate generalizes better in later iterations, and 2) there's always a probability that the expert rules are applied to avoid over-exploitation in Bayesian optimization.

For the BO surrogate, to measure its generalization ability, we define the weight as the ratio of \textbf{concordant pairs}. 
Given two configurations $x_1$ and $x_2$, the pair is concordant if the sort order of $(M(x_1),M(x_2))$ and $(y_1,y_2)$ agrees, where $M(x_1)$ is the predictive mean of $x_1$ by the BO surrogate and $y_1$ is the ground-truth performance of $x_1$. 
Since the BO surrogate is directly trained on the observations, using observations again cannot reflect its generalization ability on unseen configurations. 
Therefore, we apply cross-validation and the weight is then calculated as $w_s=\frac{2}{|D|(|D|-1)}n_s$, where $D$ is the current observations, and $n_s$ is the number of concordant pairs. 
Denote $f$ as the mapping function that maps an observed configuration $x_i$ to its corresponding fold index as $f(x_i)$. 
$n_s$ is calculated as,
\begin{equation}
\small
\begin{aligned}
    n_s=\sum_{j=1}^{\left|D\right|} \sum_{k=j+1}^{\left|D\right|} \mathds{1}( (M_{-f(x_j)}(x_j)&<M_{-f(x_k)}(x_k))  \\
    &\left. \otimes \left(y_j<y_k\right)\right).
\end{aligned}
\label{eq:generalization}
\end{equation}
where $\otimes$ is the exclusive-nor operator, and $M_{-f(x_j)}$ refers to predictive mean of the surrogate trained on observations $D$ with the $f(x_j)$-th fold left out. 
In this way, $x_j$ is not used when generating $M_{-f(x_j)}$, thus the definition is able to measure the generalization ability of the BO surrogate only by using the observations $D$.
Finally, during each iteration, there's a probability of $p_e=w_e/(w_e+w_s)$ that \sys applies the expert rule trees to select the next configuration instead of the BO acquisition function.

\subsection{Controlled History Transfer}
\label{sec:history_transfer}
In this subsection, we explain when and how \sys transfers knowledge from history tasks, i.e., previously evaluated tasks. 
As mentioned above, history knowledge may not be properly utilized in previous methods by simply transferring all tasks or the most similar one. 
In the following, we will answer two questions: 1) how to decide the set of history tasks which is potentially beneficial to the current task and 2) how to combine the knowledge of history tasks when optimizing the current task.

\subsubsection{Task Filtering.}
\label{sec:task_filtering}
In industrial scenarios, as the tuning requirements arrive and end continuously, the history database accumulates the tuning history of a large number of history tasks.
While it is almost impossible to ensure that all history tasks are similar to the current task, it is essential to filter the history tasks before optimization to ensure safe transfer.

Previous work~\cite{prats2020you} proposes an intuitive method that generates a meta-feature for each task based on the outputs from SparkEventLog and defines the distance between two tasks as the Euclidean distance between two vectors. 
Then, the most similar history task with the lowest distance is selected to aid the optimization of the current task.
However, the meta-features are often of significantly different scales, where the Euclidean distance is dominated by those dimensions with large value ranges. 
In addition, more complex relationships may hide in the meta-feature space, e.g., tasks with quite different meta-features may also be similar in tuning behavior. 

To address the above two issues, we propose to apply a regression model to learn the similarity between two given tuning tasks. 
Following \cite{prats2020you}, we vectorize each task based on 17 runtime metrics in the event log. 
The regression model takes two task meta-features as inputs and outputs their similarity, which is $(v_1, v_2) \mapsto s$, where $s \in [0,1]$ and a larger $s$ indicates the two tasks are more similar.

To train the regression model, we first generate training data based on history tasks. 
Denote the BO surrogate built on the $i^{th}$ task as $M^i$. 
Similar to Section~\ref{sec:config_gen}, for each given pair of history tasks $i$ and $j$, we define the similarity as the ratio of \textbf{concordant prediction pairs}. 
Given two configurations $x_1$ and $x_2$, the prediction pair is concordant if the sort order of $(M^i(x_1),M^i(x_2))$ and $(M^j(x_1),M^j(x_2))$ agrees, where $M^i(x_1)$ is the predictive mean of $x_1$ by the surrogate $M^i$. 
The similarity $S(i,j)$ between the $i^{th}$ and $j^{th}$ task can be computed as,
\begin{equation}
\small
\begin{aligned}
&S(i,j)=\frac{2}{|D_{r}|(|D_{r}|-1)}F(i,j), \\
&F(i,j)=\sum_{k=1}^{\left|D_{r}\right|} \sum_{l=k+1}^{\left|D_{r}\right|} \mathds{1}\left(\left(M^i(x_k)<M^i(x_l)\right)
\otimes \left(M^j(x_k)<M^j(x_l)\right) \right),
\end{aligned}
\label{eq:similarity}
\end{equation}
where $D_r$ is a set of randomly selected configurations on the search space, and $\otimes$ is the exclusive-nor operator. 
While $F(i,j)$ computes the number of concordant prediction pairs, we divide $F(i,j)$ by the number of pairs and scale it to $[0,1]$.
For each pair of history tasks $i$ and $j$, the training sample is a triple represented as $<v_i, v_j, S(i,j)>$.
To ensure symmetry, $<v_j, v_i, S(i,j)>$ is also included in the training data. 
Consider a history database with $N$ history tasks, about $N^2$ training samples are generated.
We use a Catboost regressor~\cite{dorogush2018catboost} as the machine learning model.

After training the regressor, given a new tuning task, we first generate the meta-feature of the new task and compute its similarity with all history tasks using the regressor. 
Then we filter the tasks using a threshold so that most of the irrelevant tasks are ignored.
In practice, we set a quite high threshold (0.65) to perform conservative transfer learning and avoid negative transfer caused by dissimilar history tasks.
To reduce the overhead of computation, we choose the top-5 tasks with the largest similarity as history knowledge if there are a large number of valid history tasks.
Note that, if no similar tasks are found, history transfer is disabled and the algorithm reverts to expert-assisted BO as introduced in Section~\ref{sec:expert-assisted}.

\subsubsection{Knowledge Combination.} 
\label{sec:knowledge_combination}
Given a set of similar history tasks, the next step is to combine those tuning knowledge with the BO framework.
While a simple warm-starting does not fully utilize the potential of history~\cite{feurer2015efficient}, previous methods~\cite{feurer2018scalable,zhang2021restune} propose to directly combine the predictions of different task surrogates.
However, the scale and distribution of surrogate outputs across different tasks are often different, e.g., the memory cost given the same configuration depends on the data size and running environment of a workload.
Then, the combined surrogate output in previous methods will be dominated by the surrogate where the output values are larger.
To alleviate this issue, we propose to combine the ranking output of different surrogates, as the tuning procedure only aims to find the optimum in the search space rather than getting its value.

Concretely, given a set of candidate configurations by random sampling and mutation~\cite{hutter2011sequential}, we rank the configurations based on the EI value of each task surrogate. 
We denote the ranking value of the configuration $x_i$ based on surrogate $M^j$ on the $j^{th}$ task as $R_{M^j}(x_i)$.
The combined ranking of $x_i$ is computed as,
\begin{equation}
\small
  CR(x_i)=\sum_{j=1}^{K+1}w_jR_{M^j}(x_i),
\label{eq:cr}
\end{equation}
where we assume that there are K history tasks available and the current task is the $K+1^{th}$ task. $w_j$ controls the importance of each task, which is defined as,
\begin{equation}
\small
\begin{aligned}
w_j&=\frac{exp(G(j)/ \tau)}{\sum exp(G(\cdot)/ \tau)},
    \text{where}\ \  \tau=\frac{\tau_0}{1+\text{log}\ T},\\
G(j)&=\sum_{k=1}^{\left|D\right|} \sum_{l=k+1}^{\left|D\right|} \mathds{1}\left(\left(M^j(x_k)<M^j(x_l)\right)\otimes \left(y_k<y_l\right) \right),
\end{aligned}
\end{equation}
where $D$ is the current observations, $\tau_0$ is a hyper-parameter that controls the softmax temperature $\tau$, and $T$ is the current number of BO iterations. 
$G(j)$ measures whether the surrogate $M^j$ fits the ground-truth objective values $y$.
$G(K+1)$ is obtained by cross-validation similar to Equation~\ref{eq:generalization}.
In each iteration, \sys selects the configuration with the lowest ranking value to evaluate. 
To compute the weight (Section~\ref{sec:config_gen}) for the combined surrogate when history transfer is enabled, we simply take $w_s=\sum_j w_j\frac{2*G(j)}{|D|(|D|-1)}$.
The tuning procedure will be discussed in Section~\ref{sec:tuning_procedure}.

\section{System Implementation}

\label{sec:tuning_procedure}

\begin{figure}
  \begin{center}
	\includegraphics[width=0.96\linewidth]{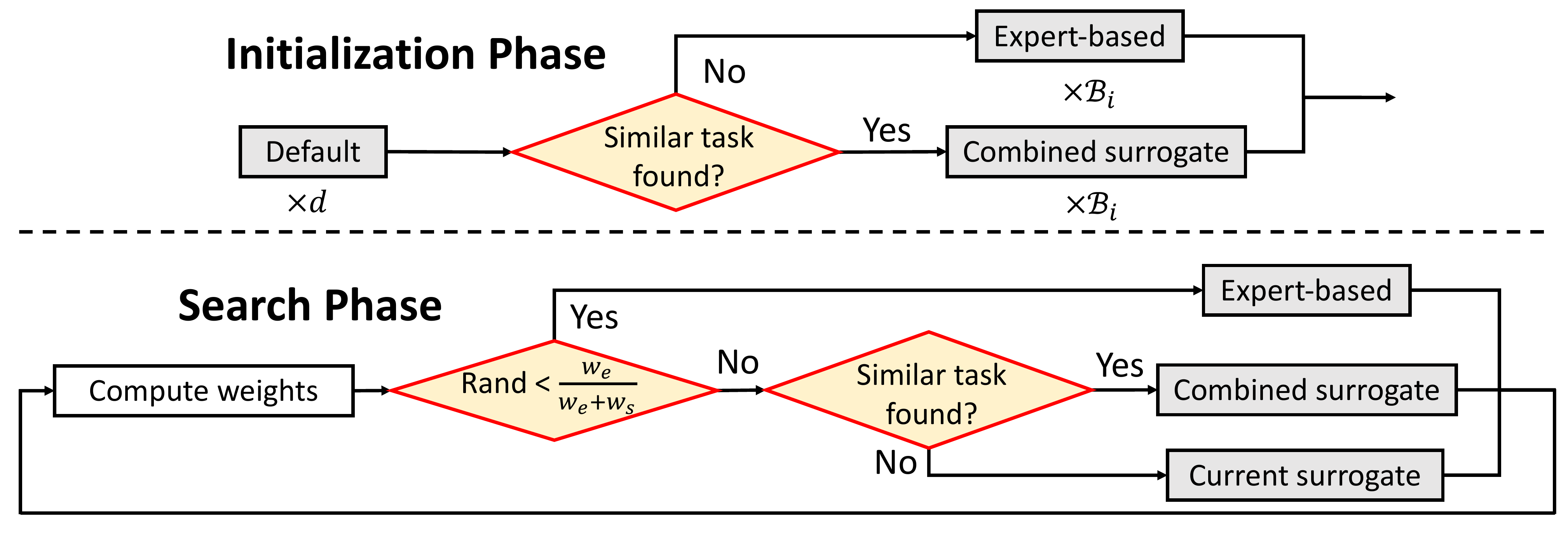}
  \end{center}
  \caption{The flow chart of the tuning procedure.}
 \label{fig:tuning_procedure}
\end{figure}

To integrate the designs introduced in Section~\ref{sec:algo}, the tuning procedure of \sys is split into two phases: the initialization and the search phase.
To make a better understanding of the tuning procedure, we also provide the flow chart of both two phases in Figure~\ref{fig:tuning_procedure}.
The pseudo-code is provided is provided in Appendix~\ref{appendix:pseudo}.

\noindent\textbf{Initialization Phase.}
During the initialization phase, \sys first evaluates 5 default configurations.
Then, it obtains the meta-feature from the running log and selects similar tasks from the history database.
If no similar tasks are found, \sys applies the expert-based initialization to generate configurations.
Else, \sys applies the combined surrogate instead until the initialization budget exhausts. 

\noindent\textbf{Search Phase.}
During the search phase, \sys first computes the weight $w_s$ for BO depending on whether there are similar history tasks and $w_e$ for expert rule trees.
Then, based on random sampling, there's a probability that the next configuration is suggested by expert knowledge.
Else, \sys samples a new configuration by optimizing the EI acquisition function or the combined ranking function when history transfer is disabled and enabled, respectively.

\noindent\textbf{Special Cases.}
When the search budget $\mathcal{B}_s$ exhausts, \sys stops the tuning procedure and returns the best-observed configuration. 
By monitoring the running information from the user interface, the end users can also early stop a tuning task if they find there's no significant improvement as expected.
In addition, as the workload data may change in online scenarios, if \sys detects a significant workload change, 
it will consider it as a new tuning task and restarts from the initialization phase. 
The demonstration of \sys functionality and user interfaces is provided in Appendix~\ref{appendix:ui}.

\section{Experimental Evaluation}
To show the practicality and effectiveness of \sys, we will investigate the following insights: 
1) Compared with state-of-the-art tuning algorithms, \sys achieves superior performance on benchmarks and real-world tasks; 
2) \sys achieves positive performance in production; 
3) The algorithm framework of \sys is more reasonable compared with other design alternatives.

\begin{figure*}[tb]
    \centering
    \includegraphics[width=0.98\linewidth]{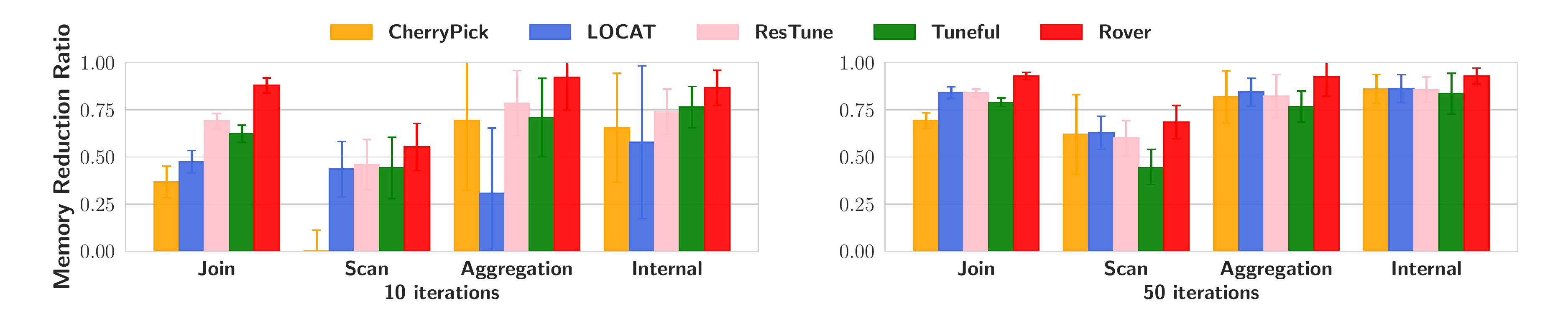}
    \caption{Memory reduction ratio (the higher, the better) with standard deviation on 3 HiBench tasks and internal tasks.}
    \label{fig:end2end}
\end{figure*}

\subsection{Setups}
\noindent\textbf{Benchmarks.}
For end-to-end comparison, we follow LOCAT~\cite{xin2022locat} and use three SQL-related tasks from the widely used Spark benchmark HiBench~\cite{huang2010hibench}:
(1) `Join' is a query that executes in two phases: Map and Reduce; 
(2) `Scan' is a query that only executes Map operations; 
(3) `Aggregation' is a query that executes the Map and Reduce operation alternately.
In addition, we also prepare 200 relatively small workloads that run internally in ByteDance, which are collected following data privacy rules.
The input sizes of workloads range from 200MB to 487GB, and the workloads involve comprehensive combinations of Map and Reduce operations.

\noindent\textbf{Baselines.}
For end-to-end comparison, we compare \sys with the following SOTA BO-based tuning algorithms: 
(1) CherryPick~\cite{alipourfard2017cherrypick}: A method with a discretized search space;
(2) Tuneful~\cite{fekry2020tuneful}: A method that explores significant parameters and applies a multi-task GP to use the most similar previous task; 
(3) LOCAT~\cite{xin2022locat}: A method that identifies important parameters and dynamically reduces the search space;
(4) ResTune~\cite{zhang2021restune}: A transfer learning method that uses all the history knowledge to accelerate the tuning process.

\noindent\textbf{Experiment Settings.}
\sys implements BO based on OpenBox~\cite{li2021openbox}, a toolkit for black-box optimization. 
The other baselines are implemented following their original papers. 
Rover uses the 10-d search space (Section~\ref{sec:search_space}), while the space for the other baselines follows the 30-d search space in Tuneful~\cite{fekry2020tuneful}.
For transfer learning, \sys takes the first 25 iterations of previous tasks as history knowledge. 
The meta-feature for each task is computed as the average runtime vectors of default configurations (Section~\ref{sec:task_filtering}).
In all experiments, we optimize the \textbf{accumulated memory cost}, which is computed as the average memory cost multiplied by the time cost (GB$\cdot$h).
We use the ratio relative to the average result of default configurations as the metric so that the result on each task is of the same scale. 
In Section~\ref{sec:analysis}, we report the mean \textbf{best-observed} result during each iteration with 5 repetitions.
The total tuning budget for each method is set to 50 iterations (5 for initialization and 45 for search).

\noindent\textbf{Environment.}
All the experiments are conducted using a virtual cluster of 24 nodes. Each node is equipped with 4 AMD EPYC 7742 64-core processors and 2T memory.
We use Spark 3.2 as our computing framework.

\subsection{End-to-end Comparison}
\label{sec:end2end}
We first provide an end-to-end comparison on three HiBench tasks and the internal tasks.
In Figure~\ref{fig:end2end}, we show the memory reduction ratio after tuning 10 and 50 iterations relative to the default configuration.
We observe that: 
1) Among Spark tuning baselines, CherryPick does not reduce the search space, thus it can not handle the tasks well. 
Though LOCAT selects important parameters dynamically during the tuning process, it requires about 15 iterations before it can shrink the space. 
Its performance is worse than \sys where a compact space is used from the beginning; 
2) Tuneful and Restune generally outperforms LOCAT in 10 iterations due to the use of similar history tasks. 
However, as they do not utilize history well, there's almost no improvement in 50 iterations; 
3) The performance of \sys is not significant on Scan, where no similar tasks are found among 202 tasks. 
Even though transfer learning is not enabled, \sys still outperforms other baselines in early and late iterations.
4) Among the competitive baselines, \sys achieves the largest memory reduction on all the tasks.
Concretely, on internal tasks, \sys further reduces \textbf{9.36\%} and \textbf{6.76\%} the memory cost relative to the baselines Tuneful and LOCAT, respectively.
We also provide results on another objective (CPU costs) in Appendix~\ref{appendix:addition_results}.

\subsection{In-production Performance}

\begin{figure}[tb]
\centering  
\subfigure[Optimization curve]{
\label{fig:industry_curve}
\includegraphics[width=0.475\linewidth]{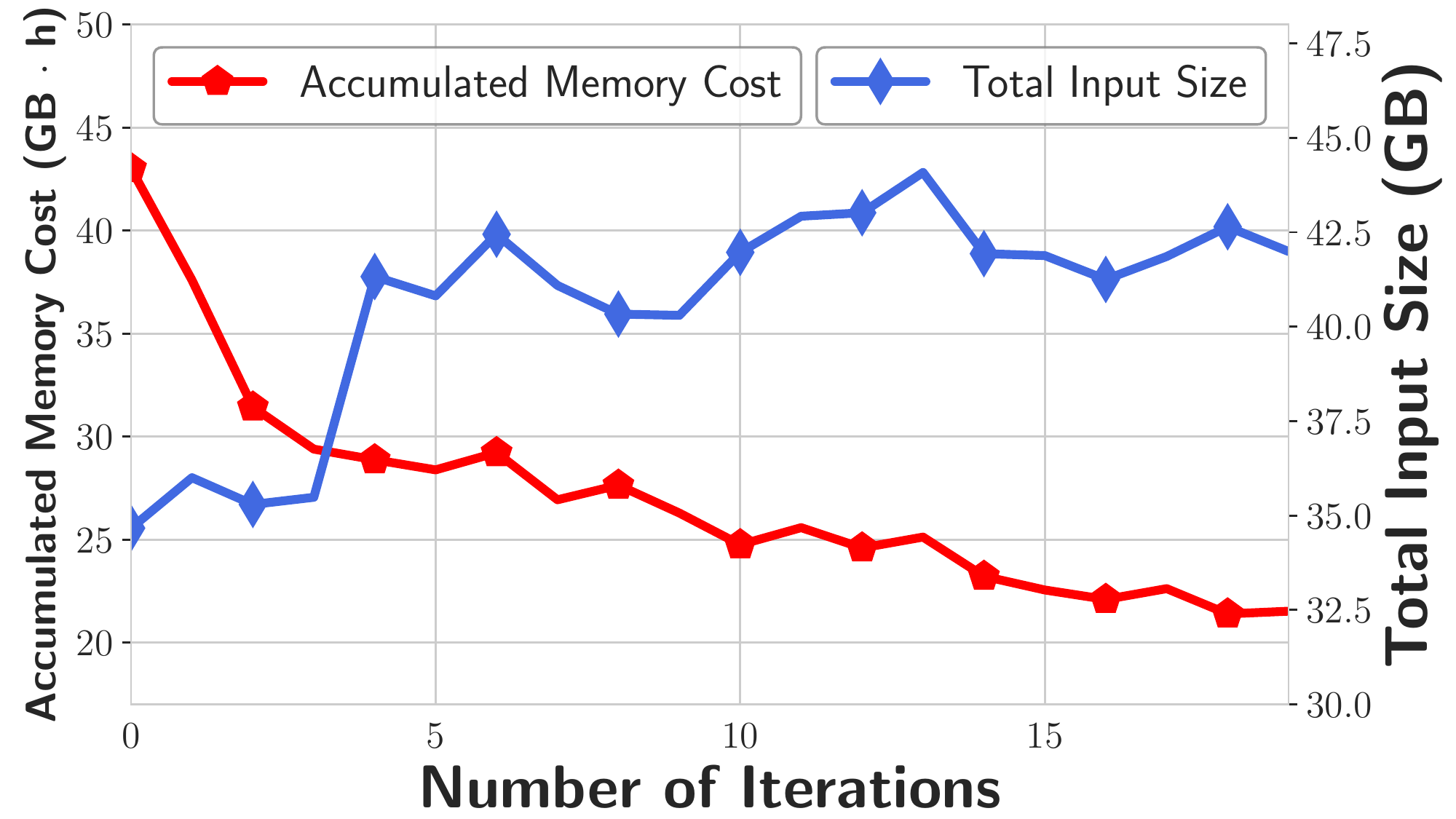}
}
\subfigure[Histogram]{
\label{fig:industry_histogram}
\includegraphics[width=0.475\linewidth]{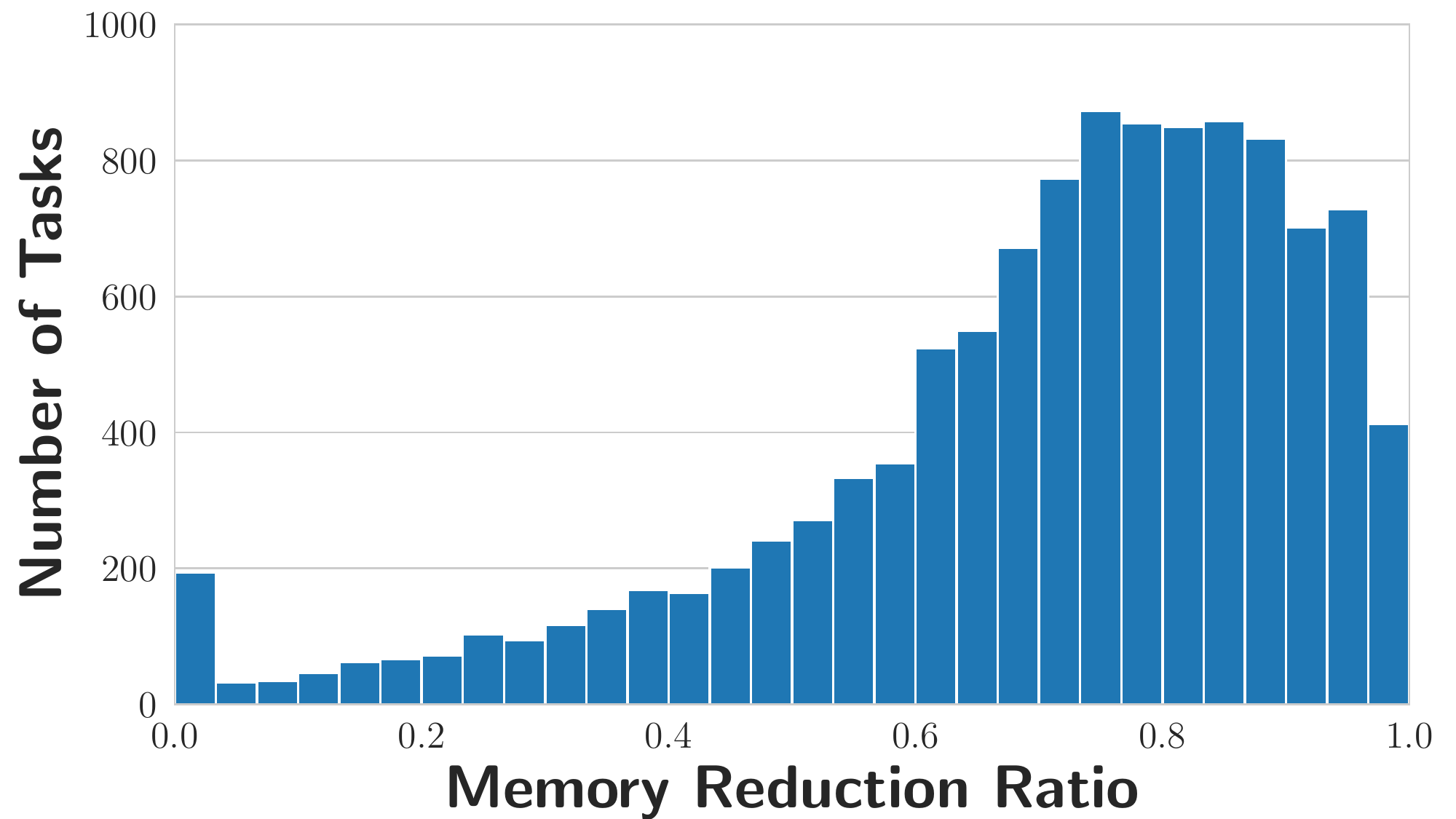}
}
\caption{In-production performance of \sys.}
\label{fig:industry}
\end{figure}

To demonstrate the practicality of \sys, we deploy the system in a production cluster and evaluate it on 12000 real-world Spark SQL tasks.
Each task is collected from actual business in ByteDance (e.g., recommendation, advertisement, etc.), which executes once a day and processes massive data generated from billions of users every day.
The input sizes of workloads range from 200MB to 37.3TB.
Since the evaluation runs in actual business and is computationally expensive, we only compare \sys with default Spark configurations suggested by Spark engineers.
The tuning budget for all tasks is set to 20 days (tuning once per day).

Figure~\ref{fig:industry_curve} shows the mean accumulated memory cost during optimization. 
We observe that \sys reduces the memory cost by 32.8\% in the initialization phase due to expert-based initialization and controlled history transfer.
After that, though the online workload gradually changes, the memory cost continuously reduces with the help of the context-aware surrogate.
The improvement is not as significant as Section~\ref{sec:end2end} because the in-production workloads are changing frequently and the default configurations for some tasks have been already tuned well by experts before optimization.
Figure~\ref{fig:industry_histogram} further shows the number of tuning tasks with different improvements, where \textbf{76.2\%} of the tasks get a significant memory reduction of over 60\%.
Moreover, \textbf{97.7\%} of the tasks get an improvement of over 10\% compared with the default configurations.
With the above improvement within 20 iterations, \sys has already saved about \textbf{\$1.1M} of the annual computing resource expense for the above 12k tasks.
While \sys currently tunes a small proportion of tasks on the platform, it is estimated that \sys may lead to a significant expense reduction when we apply it to more and larger tasks (over 320k) in near future.

\subsection{Detailed Analysis}
\label{sec:analysis}
In this part, we provide a detailed analysis of the effects of both two algorithm designs and ablation study on the algorithm framework.

\subsubsection{Analysis of Expert-assisted BO}

\begin{figure*}[htb]
\centering  
\subfigure[MSE of regressor]{
\label{fig:mse}
\includegraphics[width=0.315\linewidth]{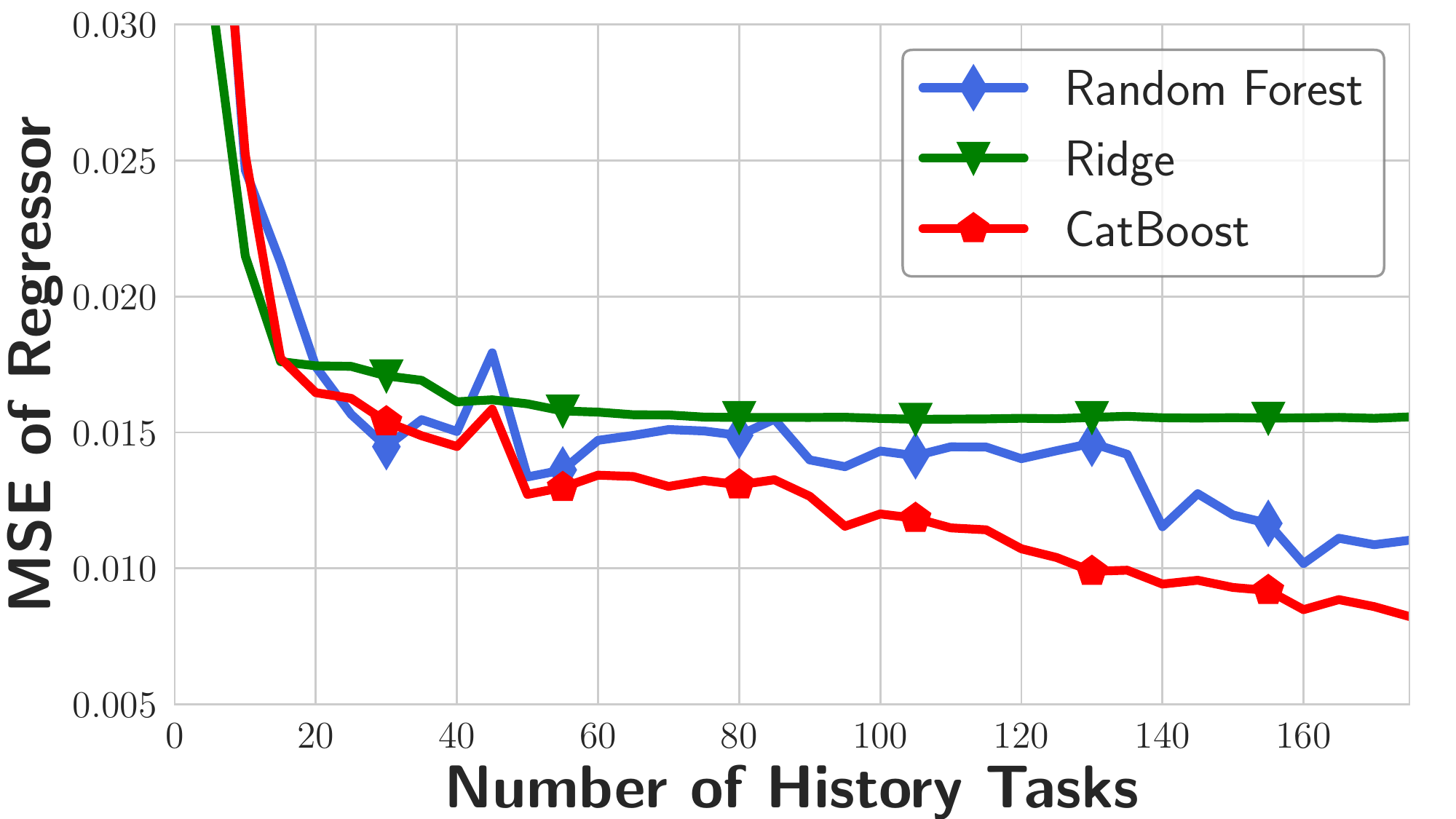}
}
\subfigure[Correlation with ground truths]{
\label{fig:correlation}
\includegraphics[width=0.315\linewidth]{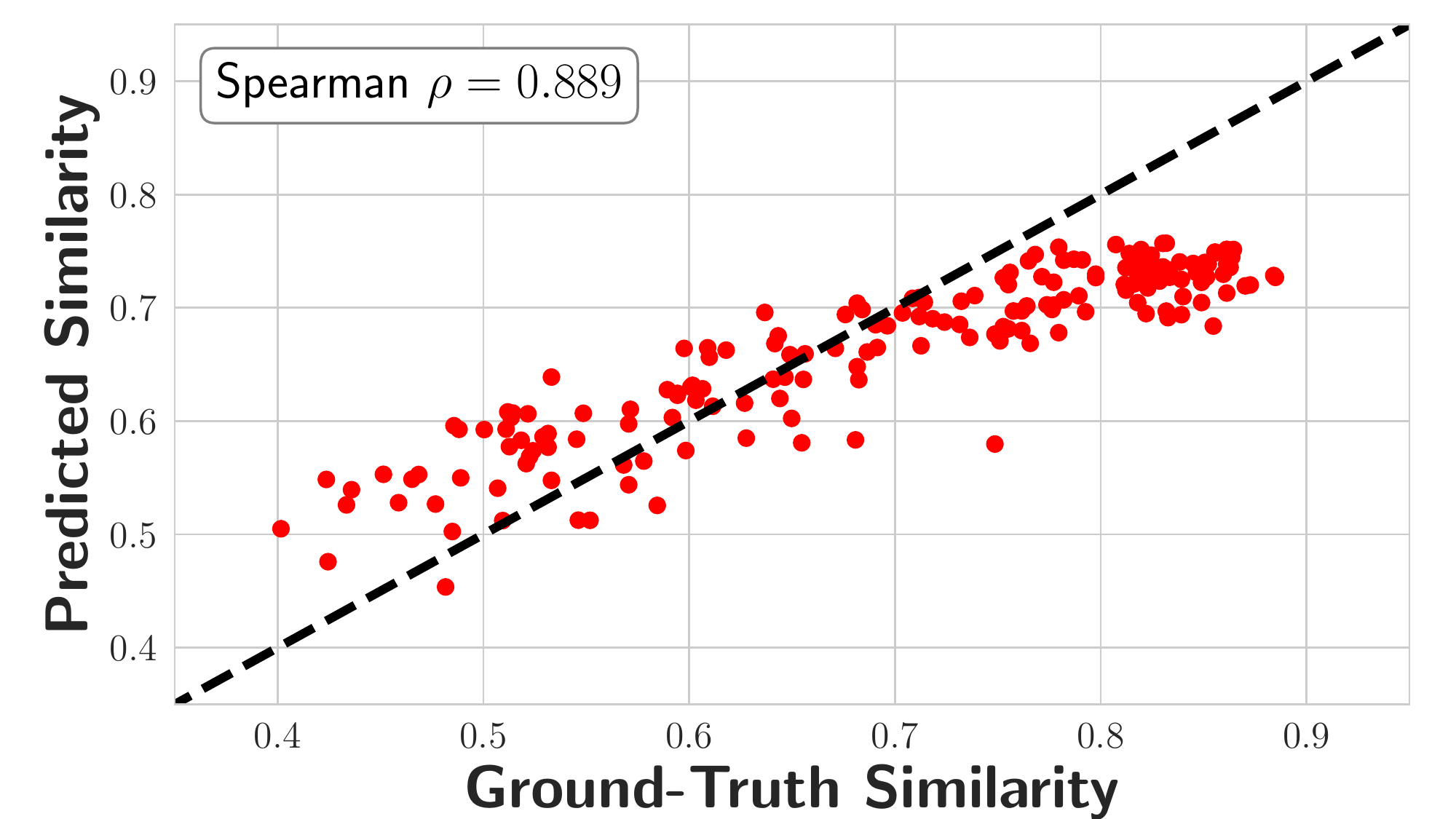}
}
\subfigure[Transfer learning strategies]{
\label{fig:transfer_learning}
\includegraphics[width=0.315\linewidth]{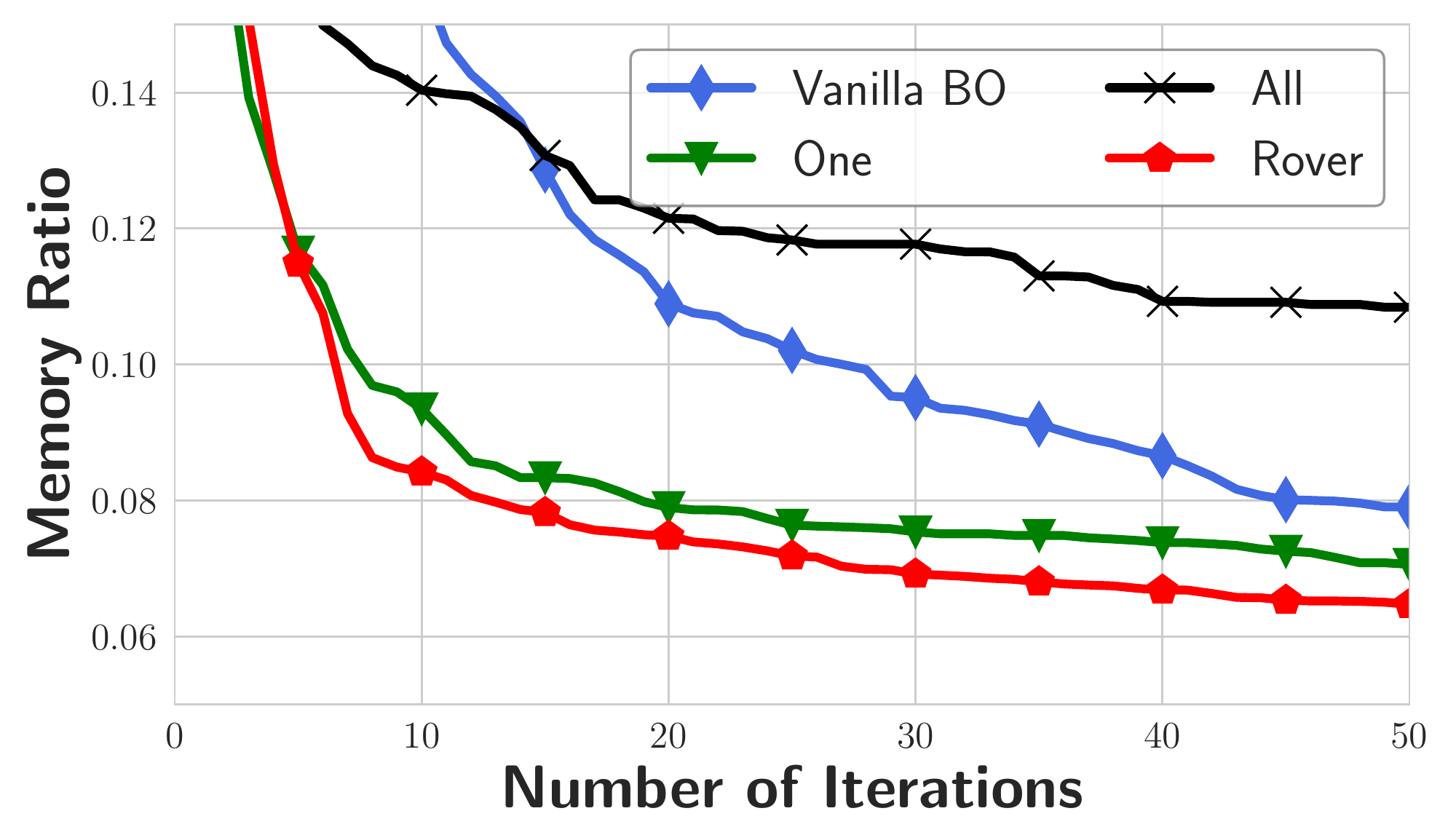}
}
\caption{Analysis of controlled history transfer.}
\label{fig:analysis_tl}
\end{figure*}

\begin{figure}[tb]
\centering  
\subfigure[Search space]{
\label{fig:search_space}
\includegraphics[width=0.475\linewidth]{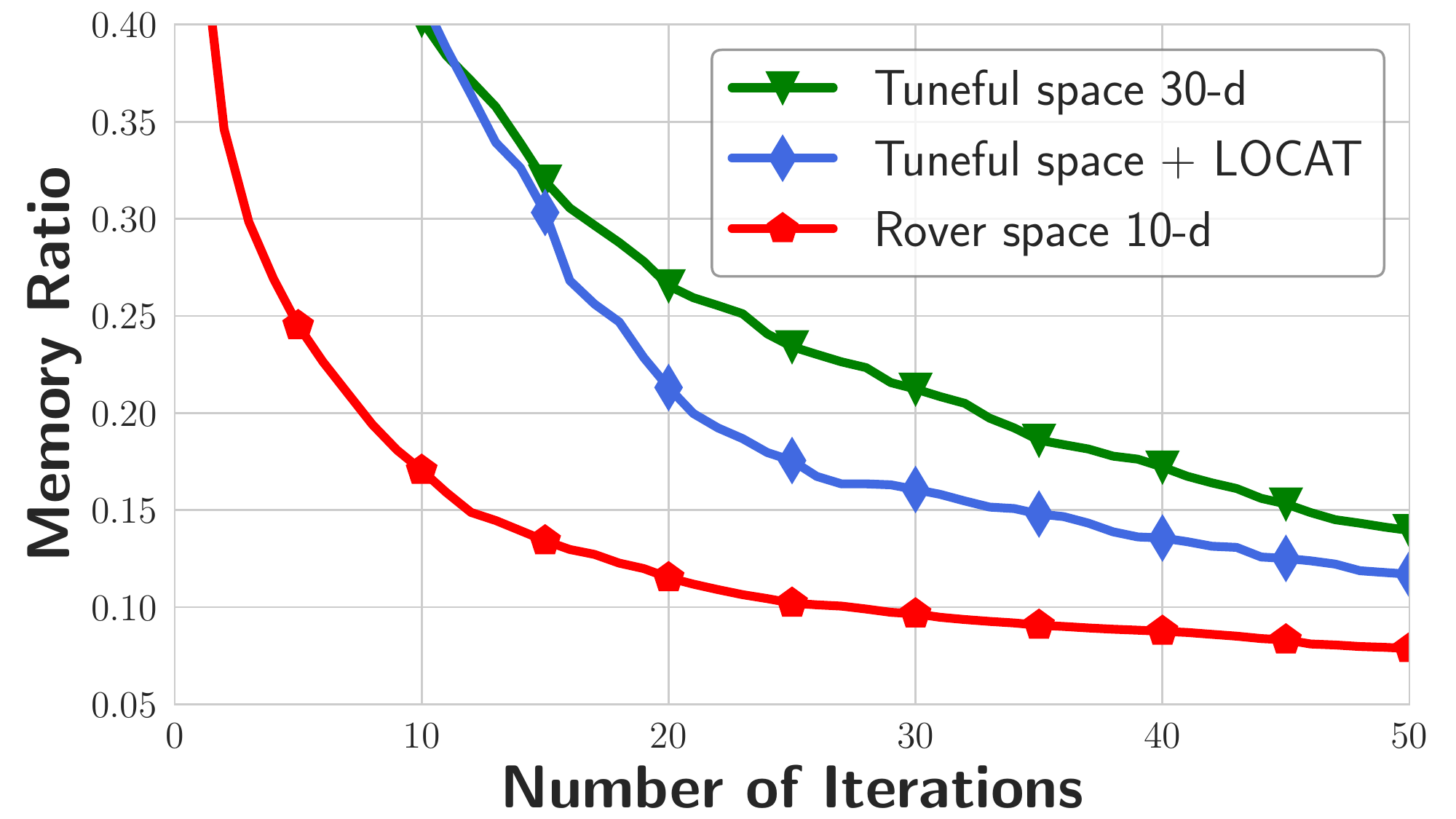}
}
\subfigure[Expert-assisted search]{
\label{fig:expert-assisted}
\includegraphics[width=0.475\linewidth]{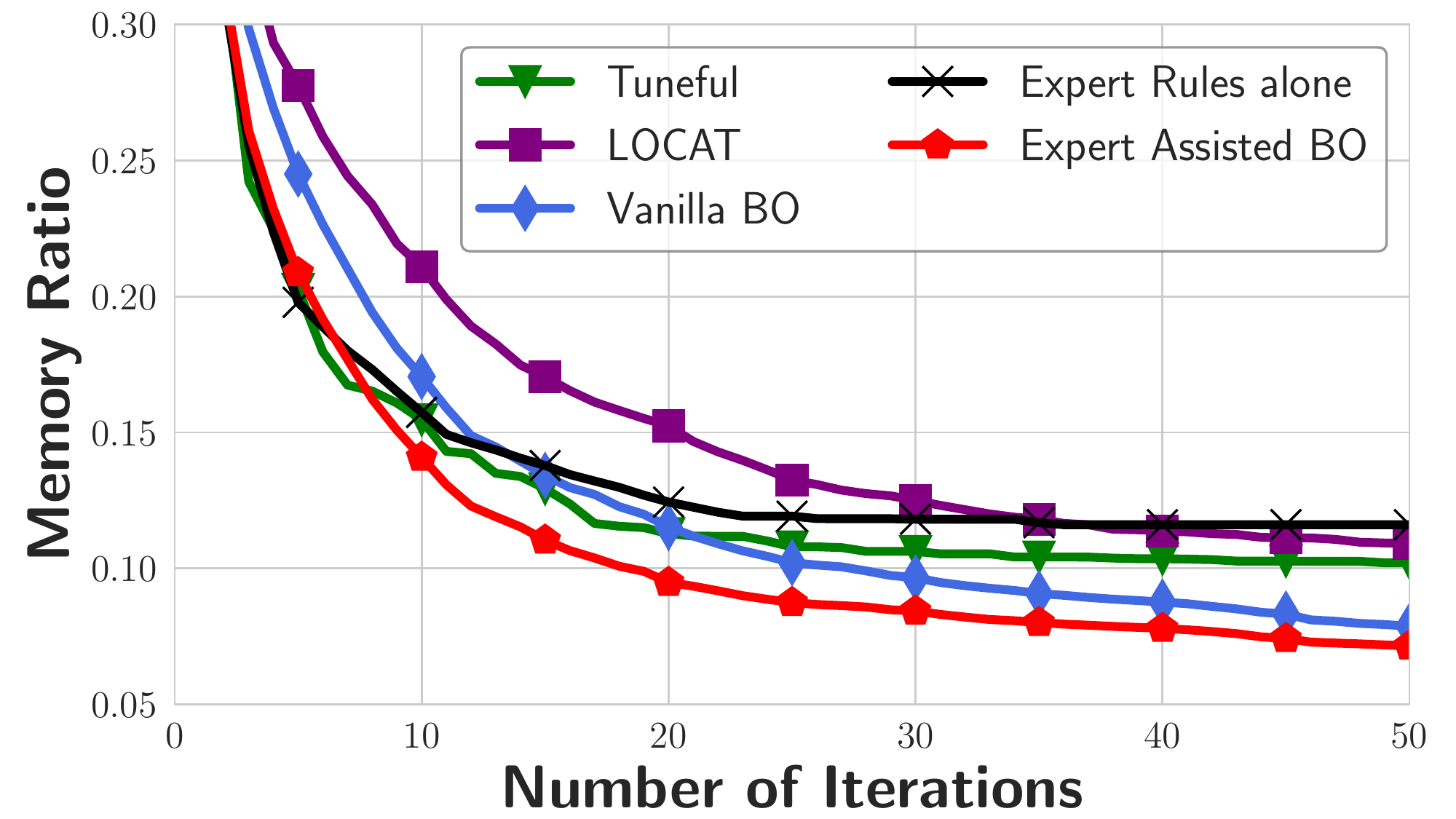}
}
\caption{Analysis of expert-assisted BO.}
\label{fig:analysis_expert}
\end{figure}

In this part, we will analyze the performance of the pre-defined search space and expert-based optimization.
We first run Bayesian optimization over three different search spaces: 1) \sys space (10 params); 2) the Tuneful~\cite{fekry2020tuneful} space (30 params); 3) the Tuneful space with dynamic shrinking used in LOCAT~\cite{xin2022locat}.
The results are shown in Figure~\ref{fig:search_space}.
We observe that the \sys space achieves clear improvement over the Tuneful space due to the use of important parameters selected by experts.
In addition, though LOCAT performs better than Tuneful, dynamic shrinking requires a number of evaluations and performs worse than BO that starts from a more compact space.
The observations show that the compact space is more practical than strategies used in state-of-the-art frameworks.

Then, we directly compare expert-based BO with other baselines using the same space in Figure~\ref{fig:expert-assisted}.
Due to initialization with expert knowledge, expert-assisted BO and expert rules achieve significant improvement over vanilla BO, LOCAT and Tuneful in the initialization phase.
After that, the memory cost of expert-assisted BO gradually decreases and consistently outperforms other baselines.
In general, at the $10^{th}$ iteration, expert-assisted BO reduces the memory cost of vanilla BO, Tuneful, and LOCAT by 2.96\%, 1.38\% and 7.00\%. 
And at the $50^{th}$ iteration, expert-assisted BO further reduces the memory cost of vanilla BO, Tuneful, and LOCAT by 0.74\%, 3.05\% and 3.80\%.
The reason that Tuneful and LOCAT performs worse than vanilla BO in our scenarios may be that 
1) multi-task BO used in Tuneful may not be effective to transfer history knowledge; 
2) space shrinking in LOCAT may discard important parameters when the space is already compact.
We also compute the average memory ratio of all configurations for expert-assisted BO and vanilla BO, which are 37.07\% and 45.43\%.
Vanilla BO leads to higher average cost because of unsafe configurations suggested by random initialization and $\epsilon$-greedy strategy.
The observations show that \sys achieves better results and performs safer than vanilla BO and using expert rules alone. 

\subsubsection{Analysis of Controller History Transfer}

We first treat 25 from 200 tasks as the test set and analyze the generalization ability of the regression model.
In Figure~\ref{fig:mse}, we plot the mean squared error (MSE) of three regression models using a different number of history tasks.
Generally, we observe that Catboost achieves the best performance compared with the other two alternatives, thus we use Catboost as the regressor.
In addition, the MSE continuously drops when the number of history tasks used for training increases.
When the number reaches 175, we also take one task from the test set and plot the predicted and ground-truth similarity in Figure~\ref{fig:correlation}.
The Spearman correlation between predictions and ground truths is 0.889, 
while the correlation between Euclidean distances and ground truths is only 0.120.
This shows that the regression model can measure the similarity of unseen tasks relatively well.

Then, we compare the controlled history transfer in \sys with other intuitive solutions: 1) \texttt{One}: transfer only the most similar task by the model; 2) \texttt{All}~\cite{zhang2021restune}: transfer all history tasks.
Since \sys only performs transfer learning when similar tasks are found, we plot the results on those tasks (43 out of 200 tasks) in Figure~\ref{fig:transfer_learning}.
We observe that \texttt{All} performs worse than vanilla BO at later iterations due to the use of too many dissimilar tasks in history.
In addition, the mean result of \sys is slightly better than \texttt{One}, but the variance of \sys is much lower (0.013 compared with 0.021 at the 50$^{th}$ iteration).
We attribute this to the use of more than one similar tasks, and the ensemble is known to produce better results and reduce variance~\cite{bian2021does,zhou2012ensemble}.
In addition, we also compare \texttt{One} with vanilla BO at the 10$^{th}$ iteration on the 157 tasks when transfer learning is \textbf{disabled} by Rover.
\texttt{One} performs worse than BO on 23\% of them, while the value is only 2 out of 43 tasks when \sys decides to use history transfer.
This shows that task filtering is essential to avoid negative transfer in production.

Finally, since history transfer is only enabled when similar tasks are found, we perform an ablation study on those tasks.
Table~\ref{tab:ablation} shows the performance gap of \sys without certain components relative to \sys.
We observe that, controlled history transfer further improves expert-based optimization about 8.42\% and 2.86\% in the first 5 and 15 iterations.
In addition, without history transfer, \sys still outperforms vanilla BO due to expert-based designs.


\begin{table}[thb]
    \centering
    \caption{Reduction ratio and gaps (\%) at different iterations relative to \sys on tasks where history transfer is enabled.}
    \resizebox{0.38\pdfpagewidth}{!}{
    \begin{tabular}{ccccccc}
    \toprule
    & 5 & $\Delta$ & 15 & $\Delta$ & 50 & $\Delta$ \\
    \midrule
    Vanilla BO on compact space & 24.39 & 13.15 & 12.78 & 5.04 & 7.86 & 1.43\\
    \sys w/o controlled history transfer & 19.66 & 8.42 & 10.60 & 2.86 & 7.32 & 0.88\\
    \midrule
    \sys & 11.24 & / & 7.74 & / & 6.44 & / \\
    \bottomrule
    \end{tabular}
    }
    \label{tab:ablation}
\end{table}

\subsubsection{Analysis of dynamic weights}
\begin{figure}[tb]
\centering  
\subfigure[Without history transfer]{
\label{fig:without_tl}
\includegraphics[width=0.475\linewidth]{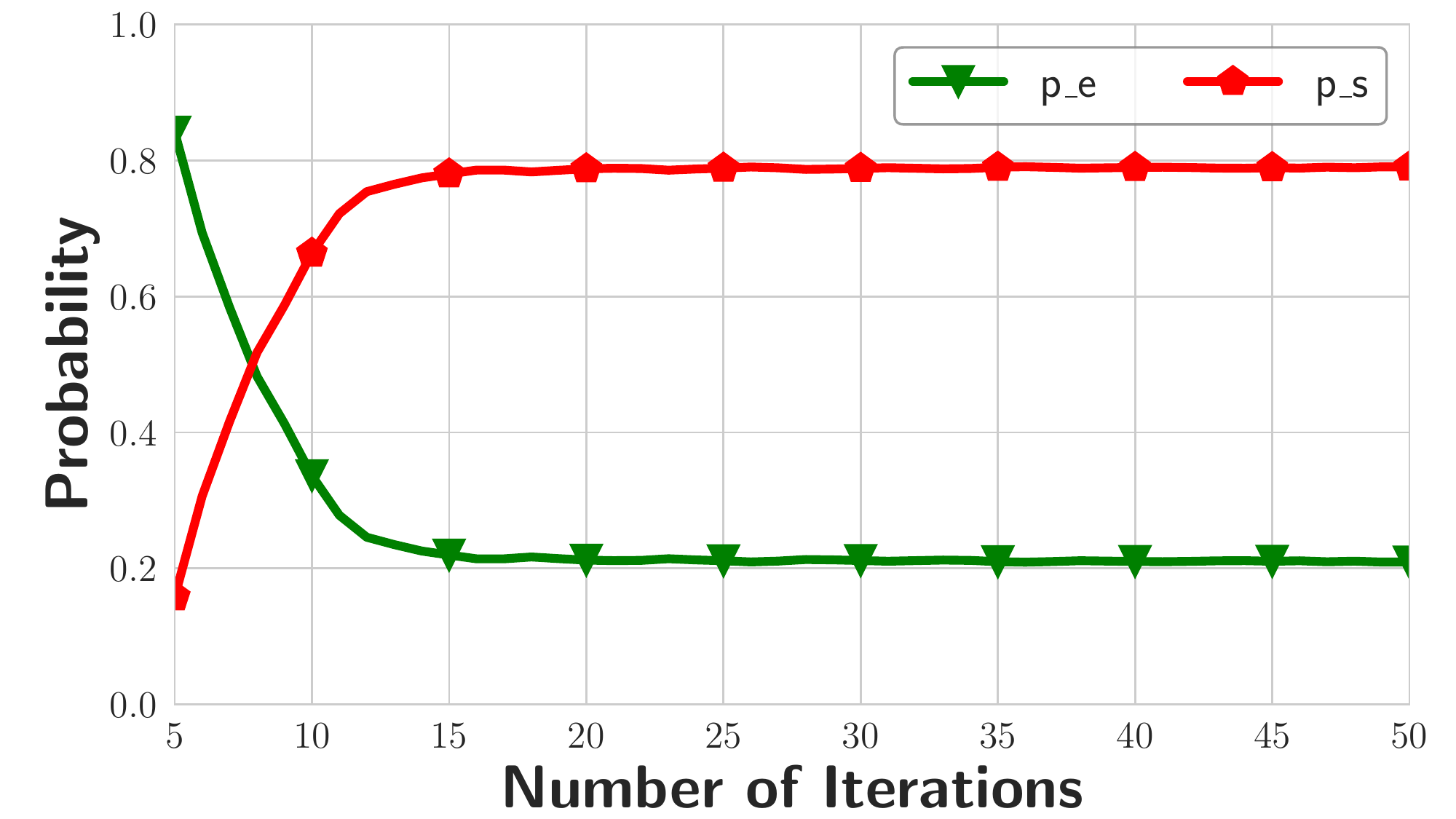}
}
\subfigure[With history transfer]{
\label{fig:with_tl}
\includegraphics[width=0.475\linewidth]{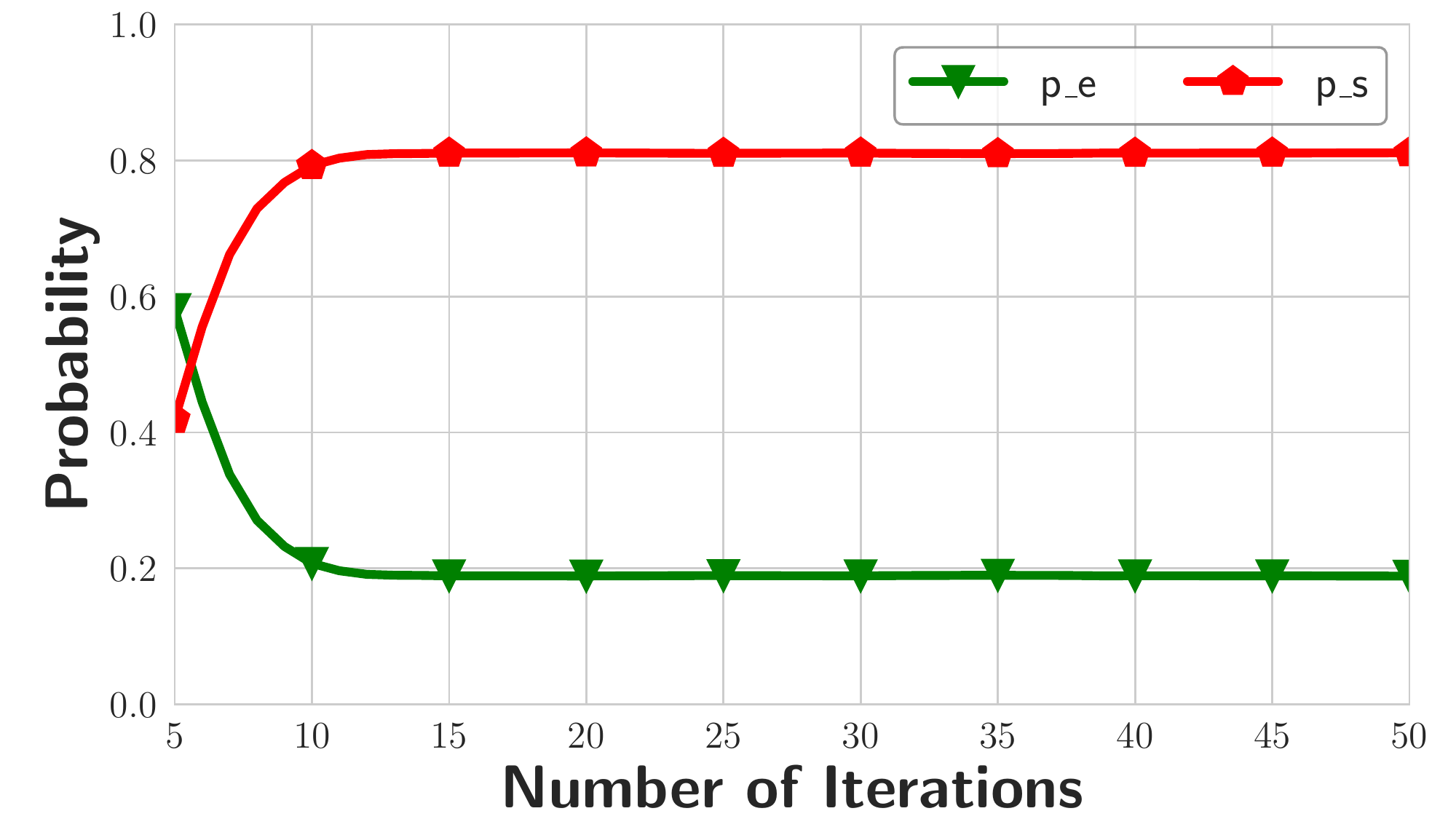}
}
\caption{Possibility of components during each iteration.}
\label{fig:analysis_weight}
\end{figure}

To show how \sys make decisions during each iteration, we show the probability of using expert rules ($p_e$) or BO ($p_s$) during optimization on 200 internal tasks in Figure~\ref{fig:analysis_weight}, which are computed as $p_e=w_e/(w_e+w_s)$ and $p_s=w_s/(w_e+w_s)$.

As mentioned in Section~\ref{sec:tuning_procedure}, we use expert rules (no similar tasks found) or surrogate ensemble (similar tasks found) in the first 5 iterations, so we display the probability after 5 iterations. 
If history transfer is disabled, the BO surrogate may be under-fitted with limited observations. 
Therefore, it takes BO another 3 iterations to get a higher probability than expert rules and it dominates the selection (over a probability of 75\%) after the $12^{th}$ iterations. 
Expert rules still have a low probability due to the constant bias as mentioned in Section~\ref{sec:config_gen}. 
If similar tasks are found, we find that the probability of BO dominates the optimization process at the $9^{th}$ iteration, which shows that the surrogate ensemble can model the relationship between configurations and performance relatively well in the beginning.

\section{Conclusion}
In this paper, we propose \sys, a deployed online Spark SQL tuning service that provides user-friendly interfaces and performs efficient and safe search on in-production workloads.
To address the re-optimization issue and enhance the tuning performance, \sys integrates expert knowledge into the search algorithm and build a strong surrogate ensemble on elaborately selected history tasks.
Extensive experiments on public benchmarks and real-world tasks show that \sys clearly outperforms competitive tuning frameworks used for Spark and databases.
Notably, when deployed in Bytedance, \sys achieves significant memory reduction on over 12k real-world Spark SQL tasks.

\nocite{hou2022dimensionality,zhu2021pre,feng2022accelerating,guo2022knowledge}

\begin{acks}
This work is supported by NSFC (No. 61832001 and U22B2037) and ByteDance-PKU joint program. Wentao Zhang and Bin Cui are the corresponding authors.
\end{acks}


\bibliographystyle{ACM-Reference-Format}
\bibliography{reference}


\clearpage
\appendix

\section{Appendix}
\subsection{Algorithm Details}
\label{appendix:pseudo}
We provide the pseudo-code of the initialization and search phase in Algorithm~\ref{algo:init} and~\ref{algo:search}, respectively.

\subsection{Discussions}
\subsubsection{Relationship with previous work}
According to~\cite{herodotou2020survey}, Rover belongs to \textbf{machine learning-based methods}. 
The main drawback of this type of method is that fitting the machine learning model often requires a large number of observations, so the optimization performance at early iterations is far from satisfactory. 
We observe the same challenge in our industrial scenario, where we need to find near-optimal configurations given limited observations. 
To tackle the challenge, Rover proposes to integrate expert rules into the machine learning-based method. 
According to the survey, using expert rules alone belongs to \textbf{rule-based methods}, which finds good configurations in the very beginning but the performance is less competitive than machine learning-based methods. 
Therefore, Rover can be briefly summarized as \textbf{a combination of rule-based methods and machine learning-based methods} to complement each other, where we improve the early performance by using expert rules and the later performance by the machine learning model. 
As shown in Section~\ref{sec:end2end}, Rover performs better than state-of-the-art machine learning-based methods like Tuneful and LOCAT.

In addition, Rover applies similar history tasks to \textbf{further improve machine learning-based methods}. The intuition is that the history surrogates are well-fitted and the surrogates of similar tasks may suggest good configurations in the current task. In Section~\ref{sec:analysis}, we also provide the ablation study to show that machine learning-based methods can be improved by both integrating expert rules and knowledge from history tasks.

\subsubsection{Advantages of Gaussian Process}
In Rover, we use the combination of Gaussian Process (GP) and Matern 5/2 kernel due to the following four reasons:

a) \textbf{Theoretical guarantee.} The theoretical characteristics of Bayesian optimization (BO) have been widely studied in previous work. For example, ~\cite{wang2018regret} and ~\cite{wang2014theoretical} provide the regret bounds of BO with Gaussian Process and Matern 5/2 kernel. ~\cite{scotto2018theoretical} studies different aspects of the convergence of BO with Gaussian Process. While Rover aims at applying BO to solve real-world data science problems, we do not make theoretical contributions to BO. Please refer to the mentioned literatures for more theoretical results.

b) \textbf{Empirical superiority.} The combination of Gaussian Process and Matern 5/2 kernel has achieved state-of-the-art in different black-box optimization problems. For example, this combination has been successfully applied in automatic algorithm hyperparameter tuning~\cite{snoek2012practical}, database knob tuning~\cite{zhang2021restune}, Spark parameter tuning~\cite{xin2022locat,fekry2020tuneful}. More specifically, an empirical paper~\cite{zhang2022facilitating} on database knob tuning compares different implementations of BO. It shows that GP+Matern achieves comparable results with SMAC and better results than TPE. In addition, the combination of GP and Matern kernel is the winning solution~\cite{jiang2021automated} of a recent competition (2021 CIKM AnalyticCup Track 2), where the algorithm is tested on synthetic functions, real-world A/B testing, and automatic algorithm hyperparameter tuning.

c) \textbf{Specialized requirements.} In our scenarios, the combination of GP and Matern kernel is more appropriate than other variants of BO. Concretely, as mentioned in Section~\ref{sec:intro}, we focus on finding good configurations with limited observations. We take TPE~\cite{bergstra2011algorithms} as a possible alternative to GP, where TPE uses kernel density estimation to model the density of configurations with good and bad performance. However, in practice, TPE requires a large number of observations for initialization, which is twice the number of parameters in the search space. Specifically, for a 30-d search space, TPE requires 60 random observations before it starts working. Therefore, we use the combination of GP and Matern kernel so that it can model the relationship between configurations and performance even with scarce observations.

d) \textbf{Fair comparison.} The last reason is that we want to make a relatively fair comparison with baselines used in Spark tuning. Tuneful and LOCAT also use the combination of GP and Matern kernel. In this way, we conclude that the improvement shown in our experiments is brought by generalized transfer learning rather than the implementation of BO surrogate.

\begin{algorithm}[t]
  \caption{Pseudo code for the initialization phase.}
  \label{algo:init}
  \begin{algorithmic}[1]
  \REQUIRE the total initialization budget $\mathcal{B}_i$, the compact search space $\mathcal{X}$ (Sec.~\ref{sec:search_space}), the default configurations $\{x_1,...,x_d\}$.
  \STATE initialize $x_{pre}=x_{d}$, observations $D=\{(x_{i},y_{i}) | i \in \{1,...,d\}\}$.
  \STATE Select similar history tasks and fetch the corresponding history surrogate set $\mathcal{M}$ (Sec.~\ref{sec:task_filtering}).
  \WHILE{$\mathcal{B}_i$ does not exhaust} 
  \IF{$\mathcal{M}=\varnothing$}
  \STATE apply the expert rule trees to $x_{pre}$ and sample $x_{new}$ from the neighborhood region (Sec.~\ref{sec:init}).
  \ELSE
  \STATE select $x_{new}$ by minimizing Equation~\ref{eq:cr} of the combined surrogate (Sec.~\ref{sec:knowledge_combination}).
  \ENDIF
  \STATE evaluate $x_{new}$ and augment $D=D\cup(x_{new},y_{new})$.
  \STATE set $x_{pre}=x_{new}$.
  \ENDWHILE
\end{algorithmic}
\end{algorithm}

\begin{algorithm}[t]
  \caption{Pseudo code for the search phase.}
  \label{algo:search}
  \begin{algorithmic}[1]
  \REQUIRE the search budget $\mathcal{B}_s$, the compact search space $\mathcal{X}$ (Sec.~\ref{sec:search_space}), the history surrogate set $\mathcal{M}$, the observations $D$.
  \ENSURE the best observed Spark configuration.
  \WHILE{$\mathcal{B}_s$ does not exhaust} 
  \IF{$\mathcal{M} \neq \varnothing$}
  \STATE Build the combined surrogate using $\mathcal{M}$ (Sec.~\ref{sec:knowledge_combination}).
  \STATE Compute the weight $w_s$ for the combined surrogate.
  \ELSE
  \STATE Compute the weight $w_s$ for the BO surrogate.
  \ENDIF
  \STATE Compute the weight $w_e$ for expert rule trees.
  \IF{$\text{rand()}<w_e/(w_e+w_s)$}
  \STATE apply the expert rule trees to $x_{pre}$ and obtain the updated configuration $x_{new}$.
  \ELSE
  \IF{$\mathcal{M} \neq \varnothing$}
  \STATE select $x_{new}$ by optimizing Equation~\ref{eq:cr}.
  \ELSE
  \STATE select $x_{new}$ by optimizing Equation~\ref{eq:ei}.
  \ENDIF
  \ENDIF
  \STATE evaluate $x_{new}$ and augment $D=D\cup(x_{new},y_{new})$.
  \STATE set $x_{pre}=x_{new}$.
  \ENDWHILE
\end{algorithmic}
\end{algorithm}

\subsubsection{Support for constrained optimization}
While \sys only supports Spark tuning without constraints so far, we discuss how to extend \sys to constrained optimization in two steps: a) how Bayesian optimization supports constraints and b) how the generalized transfer learning supports constraints.

a) As Rover does not modify the core of Bayesian optimization, constraints can be supported by applying a variant of BO using the Expected Improvement with Constraints (EIC)~\cite{gelbart2014bayesian} as the acquisition function. For problems without constraints, BO fits a performance surrogate that maps from a configuration to its estimated objective value. For problems with constraints, for each constraint, BO fits an extra constraint surrogate that maps from a configuration to the probability of satisfying this constraint. The ground-truth label is 1 for satisfying and 0 for not satisfying.

To choose the next configuration, we select the configuration with the largest expectation of the predicted improvement multiplied by the predicted probability of satisfying all constraints, which is, $\text{E}((y_{pred}-y*) \cdot \text{Pr}(c_1) \cdot ... \cdot \text{Pr}(c_n))$. Here $y_{pred}$ is the prediction of the performance surrogate, $y*$ is the best performance observed so far, and $\text{Pr}(c_i)$ is the probability prediction of the $i^{th}$ constraint surrogate. In this way, Bayesian optimization can be extended to support constraints. We refer to~\cite{gelbart2014bayesian} for more implementation details.

b) For expert rule trees, since there is a risk of suggesting configurations that violate the constraints, we can add more rules to support other targets like runtime. For example, as the parameter \texttt{sparkDaMaxExecutors} is highly related to runtime, we can simply increase its value when the runtime is large.

For controlled history transfer, to train the regressor, we define the similarity as the ratio of concordant prediction pairs of prediction under constraints. The prediction under constraints is computed as $y_{mean} \cdot \text{Pr}(c_1) \cdot ... \cdot \text{Pr}(c_n)$, where $y_{mean}$ is the mean prediction of the performance surrogate, and $\text{Pr}(c_i)$ is the probability prediction of the $i^{th}$ constraint surrogate. Then we combine different history tasks by summing up the rank of their EIC values instead of EI values.

\begin{figure*}[t]
\centering  
\subfigure[Dashboard]{
\label{fig:dashboard}
\includegraphics[width=0.485\textwidth]{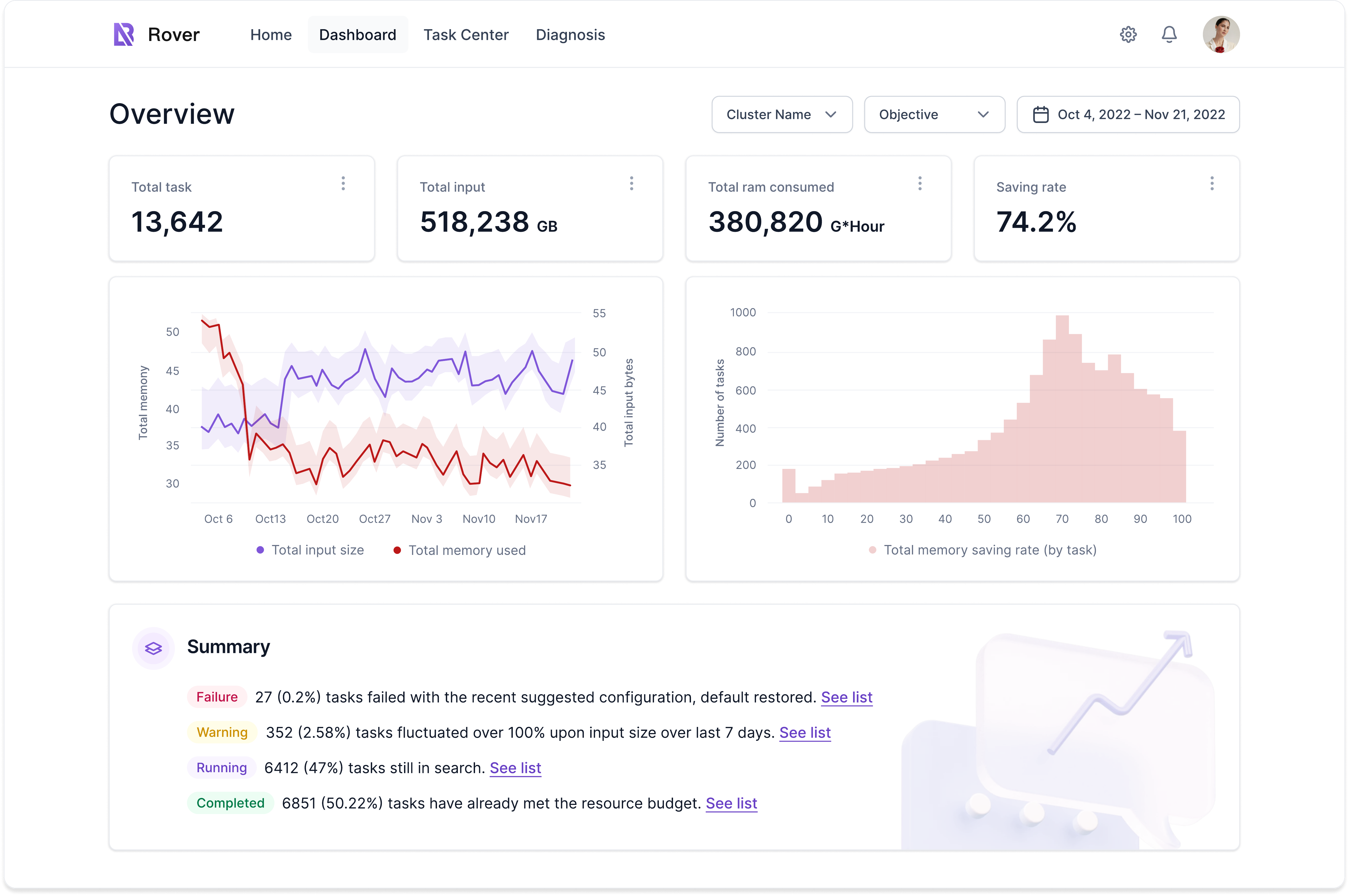}
}
\subfigure[Diagnosis Panel]{
\label{fig:diagnosis}
\includegraphics[width=0.485\textwidth]{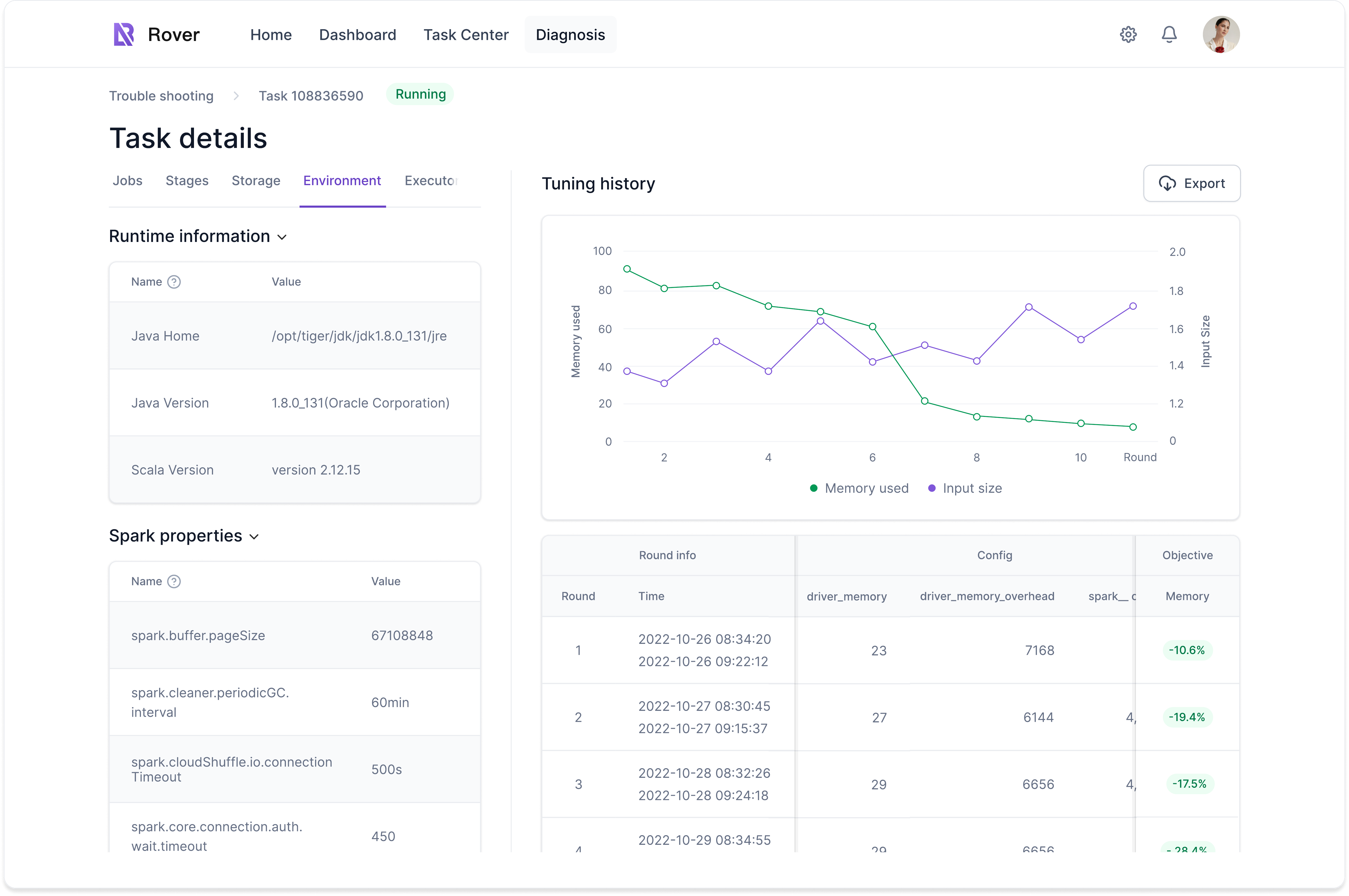}
}
\caption{Demonstration of the \sys user interface.}
\label{fig:ui}
\end{figure*}

\subsection{User Interface}
\label{appendix:ui}
We also show parts of the \sys user interfaces in Figure~\ref{fig:ui}. In Figure~\ref{fig:dashboard}, we plot the dashboard of all tuning tasks managed by the same user.
The dashboard includes two parts: 1) the overview that includes the overall cluster information, an overall optimization curve, and a histogram of the number of tasks with different improvements; 2) the summary of all tuning tasks, including reminders of failed, abnormal and finished tasks.
By clicking on a certain task in the summary list, \sys jumps to a diagnosis panel with all runtime information collected during optimization so far.
In Figure~\ref{fig:diagnosis}, we demonstrate the diagnosis panel. 
The task information (including Spark DAGs, running environment, etc.) is listed on the left side.
On the right side, \sys plots the performance curve of the chosen task and all configurations evaluated so far.
The users can then early stop the current task in the task center by analyzing the information provided by the diagnosis panel.

\subsection{Search space}
\label{appendix:space}
In Table~\ref{tab:space}, we list the 10 parameters tuned by \sys, along with their types and value ranges. Among the 10 parameters, 3 of them are categorical, which contains 3 valid values. The others are numerical, which follows a given lower and upper bound.

\begin{table}[h]
    \centering
    \caption{The 10 parameters tuned by \sys.}
    \resizebox{0.4\pdfpagewidth}{!}{
    \begin{tabular}{ccc}
        \toprule
        Name & Type & Value Range \\
        \midrule
        spark.sql.files.maxPartitionBytes & Numerical & $[16777216, 8589934592]$ \\
        spark.sql.adaptive.maxNumPostShufflePartitions & Numerical & $[20, 20000]$ \\
        spark.dynamicAllocation.maxExecutors & Numerical & $[5, 10000]$ \\
        spark.driver.cores & Categorical & $\{1, 2, 4\}$ \\
        spark.driver.memory & Numerical & $[1, 48]$ \\
        spark.driver.memoryOverhead & Numerical & $[512, 10240]$ \\
        spark.executor.cores & Categorical & $\{1, 2, 4\}$ \\
        spark.executor.memory & Numerical & $[1, 64]$ \\
        spark.executor.memoryOverhead & Numerical & $[512, 12288]$ \\
        spark.vcore.boost.ratio & Categorical & $\{1, 2, 3\}$ \\
        \bottomrule
    \end{tabular}
    }
    \label{tab:space}
\end{table}

We also provide a detailed discussion on why we optimize these 10 parameters. As mentioned in Section~\ref{sec:expert-assisted}, we calculate the SHAP importance value of 30 parameters beforehand, and the results are shown in Table~\ref{tab:shap_value}. The SHAP importance is computed based on the tuning history of 1000 internal tasks before we design Rover. Each task is tuned with over 50 iterations. The parameters in bold and italic type are selected based on SHAP importance rank and human experts, respectively.

Concretely, the top-5 parameters with the largest SHAP importance are sparkExecutorMemory, sparkExecutorMemoryOverhead, sparkExecutorCores, sparkDriverCores, and sparkDaMaxExecutors. Then we discuss why human experts choose the other five parameters.

a) sparkDriverMemory \& sparkDriverMemoryOverhead. We observe that for Spark executors, the number of cores (sparkExecutorCores), memory (sparkExecutorMemory), and memory overhead (sparkExecutorMemoryOverhead) have been already included in the search space. We also include these two parameters to perform a find-grained search for Spark drivers.

b) sparkFilesMaxPartitionBytes \& sparkAeMaxShufflePartitions. Since Map-Reduce operations are the foundation of Spark, we choose these two parameters to control the number of Mappers (sparkFilesMaxPartitionBytes) and Reducers (sparkAeMaxShufflePartitions).

c) sparkVcoreBoostRatio. This parameter is designed specifically for ByteDance infrastructure platform, which controls the number of RDD partitions a single core can process simultaneously. We suggest removing this parameter on other platforms for reproduction.

In addition, sparkDaInitialExecutors and sparkSqlAutoBroadcastJoinThreshold have relatively large SHAP importance. They are not included in the search space because,

a) sparkDaInitialExecutors. The parameter sparkDaMaxExecutors has a similar effect with sparkDaInitialExecutors. Since the former is already in the search space, we do not include the latter.

b) sparkSqlAutoBroadcastJoinThreshold. This parameter controls the threshold of table size for whether we need to store the tables in memory during Map Join. While a larger value leads to better performance, we simply set a relatively large value (40M) and do not include it in the search space.

\begin{table}[h]
    \centering
    \caption{The SHAP values of 30 parameters tuned by Tuneful.}
    \resizebox{0.4\pdfpagewidth}{!}{
    \begin{tabular}{cc}
        \toprule
        Parameter Name & SHAP Value \\
        \midrule
        \textit{sparkFilesMaxPartitionBytes}	& 1.4118541\\
\textit{sparkAeMaxShufflePartitions}&	1.161272\\
\textbf{sparkDaMaxExecutors} &	2.1909058\\
\textbf{sparkDriverCores}	&2.2878344\\
\textit{sparkDriverMemory}	&0.7766007\\
\textit{sparkDriverMemoryOverhead}	& 1.0949618\\
\textbf{sparkExecutorCores}	& 2.8089051\\
\textbf{sparkExecutorMemory}	& 9.722165\\
\textbf{sparkExecutorMemoryOverhead} &	4.940778\\
\textit{sparkVcoreBoostRatio}	& 0.8479438\\
sparkDaInitialExecutors&	1.6789923\\
sparkDaMinExecutors&	0.7285232\\
sparkAeTargetPostShuffleInputSize&	0.5051892\\
sparkAeSkewPartitionFactor	&0.6474083\\
sparkAeSkewPartitionSizeThreshold&	0.8660434\\
sparkAeBroadcastJoinThreshold&	0.38758022\\
sparkSqlAutoBroadcastJoinThreshold&	2.1695352\\
sparkSqlFilesOpenCost	&0.5295473\\
sparkSpeculationMultiplier&	0.5726126\\
sparkSpeculationQuantile	&0.2905913\\
sparkDaExecutorIdleTimeout&	0.4301605\\
sparkNetworkTimeout&	0.59897256\\
sparkShuffleIoConnectionTimeout	&0.5292596\\
sparkShuffleHighlyMapStatusThreshold	&0.34495476\\
sparkMaxRemoteBlockSizeFetchToMem	&0.64484304\\
sparkShuffleHdfsThreads&	0.34853762\\
sparkShuffleIoMaxRetries&	0.49766314\\
sparkShuffleIoRetryWait	&0.7191459\\
sparkSqlInMemoryColumnarStorageBatchSize	&0.21765669\\
sparkKryoserializerBufferMax	&0.5383134\\
        \bottomrule
    \end{tabular}
    }
    \label{tab:shap_value}
\end{table}

\subsection{Expert Rules}
\label{appendix:rules}
In Tables~\ref{tab:rule1} and ~\ref{tab:rule2}, we list all the expert rules designed for the 10 parameters.
In each table, we show the parameter name, adjustment condition, direction, step, and bounds.

\begin{table*}[htb]
    \centering
    \caption{All expert rules used in the main experiments of \sys (1/2).}
    \resizebox{0.8\pdfpagewidth}{!}{
    \begin{tabular}{ccccc}
        \toprule
        Name & Condition & Direction & Step & Bounds \\
        \midrule
        spark.sql.files.maxPartitionBytes &  stage\_max\_avg\_input\_run\_time $\leq$ 0.2 & $\uparrow$ & *2 & Lower: 16M, Upper: 4G\\
        \hline
        spark.sql.files.maxPartitionBytes & stage\_max\_avg\_input\_run\_time $\geq$ 0.4 & $\downarrow$ & *0.5 & Lower: 16M, Upper: 4G\\
        \midrule
        \multirow{5}{*}{spark.vcore.boost.ratio} & ifnull(spark.vcore.boost.ratio, 1) $\leq$ 2 & \multirow{5}{*}{$\uparrow$} & \multirow{5}{*}{/} & \multirow{5}{*}{Lower:3, Upper:3}\\
        & and ((0 < max\_mem\_usage $\leq$ 0.55  & & & \\
        & and 0 < avg\_mem\_usage $\leq$ 0.45) & & & \\
        & or avg\_mem\_usage = 0) & & & \\
        & and stage\_max\_avg\_tasks\_run\_time $\leq$ 0.2 & & & \\
        \hline
        \multirow{4}{*}{spark.vcore.boost.ratio} & spark.vcore.boost.ratio = 3 & \multirow{4}{*}{$\downarrow$} & \multirow{4}{*}{/} & \multirow{4}{*}{Lower:2, Upper:2}\\
        & and (max\_mem\_usage $\geq$ 0.30   & & & \\
        & or avg\_mem\_usage = 0) & & & \\
        & and stage\_max\_avg\_tasks\_run\_time $\geq$ 0.4 & & & \\
        \midrule
        \multirow{2}{*}{spark.executor.cores} & spark.executor.cores = 2  & \multirow{2}{*}{$\downarrow$} & \multirow{2}{*}{/} & \multirow{2}{*}{Lower:1, Upper:1}\\
        & and stage\_max\_avg\_tasks\_run\_time $\leq$ 0.1667  & & & \\
        \hline
        \multirow{4}{*}{spark.executor.cores} & spark.executor.cores = 1 & \multirow{4}{*}{$\uparrow$} & \multirow{4}{*}{/} & \multirow{4}{*}{Lower:2, Upper:2}\\
        & and stage\_max\_avg\_tasks\_run\_time $\geq$ 0.3334  & & & \\
        & and not(0.90 $\leq$ max\_mem\_usage < 2.0   & & & \\
        & and 0.74 $\leq$ avg\_mem\_usage < 2.0)  & & & \\
        \hline
        \multirow{4}{*}{spark.executor.cores} & spark.executor.cores = 1 & \multirow{4}{*}{$\uparrow$} & \multirow{4}{*}{/} & \multirow{4}{*}{Lower:2, Upper:2}\\
        & and stage\_max\_avg\_tasks\_run\_time $\geq$ 0.3334  & & & \\
        & and 0.90 $\leq$ max\_mem\_usage < 2.0   & & & \\
        & and 0.74 $\leq$ avg\_mem\_usage < 2.0  & & & \\
        \midrule
        \multirow{2}{*}{spark.executor.memory} & spark.executor.cores = 2 & \multirow{2}{*}{$\downarrow$} & \multirow{2}{*}{*0.5} & \multirow{2}{*}{Lower:1, Upper:64}\\
        & and stage\_max\_avg\_tasks\_run\_time $\leq$ 0.1667  & & & \\
        \hline
        \multirow{3}{*}{spark.executor.memory} & spark.executor.cores = 2 & \multirow{3}{*}{$\downarrow$} & \multirow{3}{*}{*0.9}  & \multirow{3}{*}{Lower:1, Upper:64}\\
        & and 0 < max\_mem\_usage $\leq$ 0.72  & &  & \\
        & and 0 < avg\_mem\_usage $\leq$ 0.60  & &  & \\
        \hline
        \multirow{3}{*}{spark.executor.memory} & spark.executor.cores = 1 & \multirow{3}{*}{$\downarrow$} & \multirow{3}{*}{*0.9}  & \multirow{3}{*}{Lower:1, Upper:64}\\
        & and 0 < max\_mem\_usage $\leq$ 0.72  & & & \\
        & and 0 < avg\_mem\_usage $\leq$ 0.60  & & & \\
        \hline
        \multirow{3}{*}{spark.executor.memory} & spark.executor.cores = 2 & \multirow{3}{*}{$\uparrow$} & \multirow{3}{*}{*1.1}  & \multirow{3}{*}{Lower:1, Upper:64}\\
        & and 0.90 $\leq$ max\_mem\_usage < 2.0 & & & \\
        & and 0.74 $\leq$ avg\_mem\_usage < 2.0 & & & \\
        \hline
        \multirow{3}{*}{spark.executor.memory} & spark.executor.cores = 1 & \multirow{3}{*}{$\uparrow$} & \multirow{3}{*}{*1.1} & \multirow{3}{*}{Lower:1, Upper:64}\\
        & and 0.90 $\leq$ max\_mem\_usage < 2.0 & & & \\
        & and 0.74 $\leq$ avg\_mem\_usage < 2.0 & & & \\
        \hline
        \multirow{4}{*}{spark.executor.memory} & spark.executor.cores = 1 & \multirow{4}{*}{$\uparrow$} & \multirow{4}{*}{*2} & \multirow{4}{*}{Lower:1, Upper:64}\\
        & and stage\_max\_avg\_tasks\_run\_time >= 0.3334  & &  & \\
        & and not (0.90 $\leq$ max\_mem\_usage < 2.0   & & & \\
        & and 0.74 $\leq$ avg\_mem\_usage < 2.0)   & &  & \\
        \hline
        \multirow{4}{*}{spark.executor.memory} & spark.executor.cores = 1 & \multirow{4}{*}{$\uparrow$} & spark.executor.memory = & \multirow{4}{*}{Lower:1, Upper:64}\\
        & and stage\_max\_avg\_tasks\_run\_time >= 0.3334  & & spark.executor.memory * 2.2 + & \\
        & and 0.90 $\leq$ max\_mem\_usage < 2.0  & &  spark.executor.memoryOverhead / 1024 & \\
        & and 0.74 $\leq$ avg\_mem\_usage < 2.0   & &  & \\
        \midrule
        \multirow{2}{*}{spark.executor.memoryOverhead} & spark.executor.cores = 2 & \multirow{2}{*}{$\downarrow$} & \multirow{2}{*}{*0.5} & \multirow{2}{*}{Lower:512, Upper:12K}\\
        & and stage\_max\_avg\_tasks\_run\_time $\leq$ 0.1667  & & & \\
        \hline
        \multirow{3}{*}{spark.executor.memoryOverhead} & spark.executor.cores = 2 & \multirow{3}{*}{$\downarrow$} & \multirow{3}{*}{*0.9}  & \multirow{3}{*}{Lower:512, Upper:12K}\\
        & and 0 < max\_mem\_usage $\leq$ 0.72  & &  & \\
        & and 0 < avg\_mem\_usage $\leq$ 0.60  & & & \\
        \hline
        \multirow{3}{*}{spark.executor.memoryOverhead} & spark.executor.cores = 1 & \multirow{3}{*}{$\downarrow$} & \multirow{3}{*}{*0.9} & \multirow{3}{*}{Lower:512, Upper:12K}\\
        & and 0 < max\_mem\_usage $\leq$ 0.72  & &  & \\
        & and 0 < avg\_mem\_usage $\leq$ 0.60  & &  & \\
        \hline
        \multirow{3}{*}{spark.executor.memoryOverhead} & spark.executor.cores = 2 & \multirow{3}{*}{$\uparrow$} & \multirow{3}{*}{*1.1}  & \multirow{3}{*}{Lower:512, Upper:12K}\\
        & and 0.90 $\leq$ max\_mem\_usage < 2.0  & &  & \\
        & and 0.74 $\leq$ avg\_mem\_usage < 2.0  & &  & \\
        \hline
        \multirow{3}{*}{spark.executor.memoryOverhead} & spark.executor.cores = 1 & \multirow{3}{*}{$\uparrow$} & \multirow{3}{*}{*1.1} & \multirow{3}{*}{Lower:512, Upper:12K}\\
        & and 0.90 $\leq$ max\_mem\_usage < 2.0  & &  & \\
        & and 0.74 $\leq$ avg\_mem\_usage < 2.0  & &  & \\
        \hline
        \multirow{4}{*}{spark.executor.memoryOverhead} & spark.executor.cores = 1 & \multirow{4}{*}{$\uparrow$} & \multirow{4}{*}{*2} & \multirow{4}{*}{Lower:512, Upper:12K}\\
        & and stage\_max\_avg\_tasks\_run\_time >= 0.3334  & &  & \\
        & and not (0.90 $\leq$ max\_mem\_usage < 2.0   & & & \\
        & and 0.74 $\leq$ avg\_mem\_usage < 2.0)   & &  & \\
        \hline
        \multirow{4}{*}{spark.executor.memoryOverhead} & spark.executor.cores = 1 & \multirow{4}{*}{$\uparrow$} & \multirow{4}{*}{*2.2} & \multirow{4}{*}{Lower:512, Upper:12K}\\
        & and stage\_max\_avg\_tasks\_run\_time >= 0.3334  & &  & \\
        & and 0.90 $\leq$ max\_mem\_usage < 2.0   & &  & \\
        & and 0.74 $\leq$ avg\_mem\_usage < 2.0   & &  & \\
        \bottomrule
    \end{tabular}
    }
    
    \label{tab:rule1}
\end{table*}

\begin{table*}[htb]
    \centering
    \caption{All expert rules used in the main experiments of \sys (2/2).}
    \resizebox{0.8\pdfpagewidth}{!}{
    \begin{tabular}{ccccc}
        \toprule
        Name & Condition & Direction & Step & Bounds \\
        \midrule
        spark.dynamicAllocation.maxExecutors  &  0 $\leq$ total\_memory $\leq$ 3 & / & / & Lower: 5, Upper: 5\\
        \hline
        spark.dynamicAllocation.maxExecutors  &  3 < total\_memory $\leq$ 5 & / & / & Lower: 6, Upper: 6\\
        \hline
        spark.dynamicAllocation.maxExecutors  &  5 < total\_memory $\leq$ 8 & / & / & Lower: 8, Upper: 8\\
        \hline
        spark.dynamicAllocation.maxExecutors  &  8 < total\_memory $\leq$ 15 & / & / & Lower: 10, Upper: 10\\
        \hline
        spark.dynamicAllocation.maxExecutors  &  15 < total\_memory $\leq$ 20 & / & / & Lower: 15, Upper: 15\\
        \hline
        spark.dynamicAllocation.maxExecutors  &  20 < total\_memory $\leq$ 40 & / & / & Lower: 20, Upper: 20\\
        \midrule
        \multirow{3}{*}{spark.sql.adaptive.maxNumPostShufflePartitions} & spark.sql.adaptive.enabled = 'true' & \multirow{3}{*}{$\downarrow$} & \multirow{3}{*}{*0.5} & \multirow{3}{*}{Lower:40, Upper:400}\\
        & and spark.sql.adaptive.maxNumPostShufflePartitions $\geq$ 40   & &  & \\
        & and 0 $\leq$ stage\_max\_avg\_shuffle\_read\_run\_time $\leq$ 0.1667  & &  & \\
        \hline
        \multirow{2}{*}{spark.sql.adaptive.maxNumPostShufflePartitions} & spark.sql.adaptive.enabled = 'true' & \multirow{2}{*}{$\uparrow$} & \multirow{2}{*}{*2} & \multirow{2}{*}{Lower:80, Upper:800}\\
        & and stage\_max\_avg\_shuffle\_read\_run\_time $\geq$ 1.2 & &  & \\
        \midrule
        \multirow{2}{*}{spark.driver.cores} & 0 < max\_driver\_mem\_usage $\leq$ 0.35  & \multirow{2}{*}{/} & \multirow{2}{*}{/} & \multirow{2}{*}{Lower:1, Upper:1}\\
        & and 0 < avg\_driver\_mem\_usage $\leq$ 0.25 & &  & \\
        \hline
        \multirow{2}{*}{spark.driver.cores} & 0 < max\_driver\_mem\_usage $\leq$ 0.45  & \multirow{2}{*}{/} & \multirow{2}{*}{/} & \multirow{2}{*}{Lower:1, Upper:1}\\
        & and 0 < avg\_driver\_mem\_usage $\leq$ 0.35 & &  & \\
        \hline
        \multirow{2}{*}{spark.driver.cores} & 0 < max\_driver\_mem\_usage $\leq$ 0.60  & \multirow{2}{*}{/} & \multirow{2}{*}{/} & \multirow{2}{*}{Lower:1, Upper:1}\\
        & and 0 < avg\_driver\_mem\_usage $\leq$ 0.50 & &  & \\
        \hline
        \multirow{2}{*}{spark.driver.cores} & 0.92 $\leq$ max\_driver\_mem\_usage $\leq$ 2.0  & \multirow{2}{*}{/} & \multirow{2}{*}{/} & \multirow{2}{*}{Lower:1, Upper:1}\\
        & and 0.92 $\leq$ avg\_driver\_mem\_usage $\leq$ 2.0 & &  & \\
        \hline
        \multirow{2}{*}{spark.driver.cores} & 0.8 $\leq$ max\_driver\_mem\_usage $\leq$ 1.0  & \multirow{2}{*}{/} & \multirow{2}{*}{/} & \multirow{2}{*}{Lower:1, Upper:1}\\
        & and 0.72 $\leq$ avg\_driver\_mem\_usage $\leq$ 1.0 & &  & \\
        \hline
        \multirow{2}{*}{spark.driver.cores} & 0.9 $\leq$ max\_driver\_mem\_usage $\leq$ 1.0  & \multirow{2}{*}{/} & \multirow{2}{*}{/} & \multirow{2}{*}{Lower:1, Upper:1}\\
        & and 0.6 $\leq$ avg\_driver\_mem\_usage $\leq$ 1.0 & &  & \\
        \midrule
        \multirow{3}{*}{spark.driver.memory} & 0 < max\_driver\_mem\_usage $\leq$ 0.35  & \multirow{3}{*}{$\downarrow$} & \multirow{3}{*}{*0.8} & \multirow{3}{*}{Lower:1, Upper:48}\\
        & and 0 < avg\_driver\_mem\_usage $\leq$ 0.25 & & & \\
        & & & & \\
        \hline
        \multirow{3}{*}{spark.driver.memory} & 0 < max\_driver\_mem\_usage $\leq$ 0.45  & \multirow{3}{*}{$\downarrow$} & \multirow{3}{*}{*0.85} & \multirow{3}{*}{Lower:1, Upper:48}\\
        & and 0 < avg\_driver\_mem\_usage $\leq$ 0.35 & & & \\
        & & & & \\
        \hline
        \multirow{3}{*}{spark.driver.memory} & 0 < max\_driver\_mem\_usage $\leq$ 0.60  & \multirow{3}{*}{$\downarrow$} & \multirow{3}{*}{*0.9} & \multirow{3}{*}{Lower:1, Upper:48}\\
        & and 0 < avg\_driver\_mem\_usage $\leq$ 0.50 & & & \\
        & & &  & \\
        \hline
        \multirow{3}{*}{spark.driver.memory} & 0.92 $\leq$ max\_driver\_mem\_usage $\leq$ 2.0  & \multirow{3}{*}{$\uparrow$} & \multirow{3}{*}{*1.2} & \multirow{3}{*}{Lower:1, Upper:48}\\
        & and 0.92 $\leq$ avg\_driver\_mem\_usage $\leq$ 2.0 & & & \\
        & & &  & \\
        \hline
        \multirow{3}{*}{spark.driver.memory} & 0.8 $\leq$ max\_driver\_mem\_usage $\leq$ 1.0  & \multirow{3}{*}{$\uparrow$} & \multirow{3}{*}{*1.1} & \multirow{3}{*}{Lower:1, Upper:48}\\
        & and 0.72 $\leq$ avg\_driver\_mem\_usage $\leq$ 1.0 & & & \\
        & & & & \\
        \hline
        \multirow{3}{*}{spark.driver.memory} & 0.9 $\leq$ max\_driver\_mem\_usage $\leq$ 1.0  & \multirow{3}{*}{$\uparrow$} & \multirow{3}{*}{*1.1} & \multirow{3}{*}{Lower:1, Upper:48}\\
        & and 0.6 $\leq$ avg\_driver\_mem\_usage $\leq$ 1.0 & & & \\
        & & & & \\
        \midrule
        \multirow{2}{*}{spark.driver.memoryOverhead} & 0 < max\_driver\_mem\_usage $\leq$ 0.35  & \multirow{2}{*}{$\downarrow$} & \multirow{2}{*}{*0.8} & \multirow{2}{*}{Lower:512, Upper:10K}\\
        & and 0 < avg\_driver\_mem\_usage $\leq$ 0.25 & &  & \\
        \hline
        \multirow{2}{*}{spark.driver.memoryOverhead} & 0 < max\_driver\_mem\_usage $\leq$ 0.45  & \multirow{2}{*}{$\downarrow$} & \multirow{2}{*}{*0.85} & \multirow{2}{*}{Lower:512, Upper:10K}\\
        & and 0 < avg\_driver\_mem\_usage $\leq$ 0.35 & &  & \\
        \hline
        \multirow{2}{*}{spark.driver.memoryOverhead} & 0 < max\_driver\_mem\_usage $\leq$ 0.60  & \multirow{2}{*}{$\downarrow$} & \multirow{2}{*}{*0.9} & \multirow{2}{*}{Lower:512, Upper:10K}\\
        & and 0 < avg\_driver\_mem\_usage $\leq$ 0.50 & &  & \\
        \hline
        \multirow{2}{*}{spark.driver.memoryOverhead} & 0.92 $\leq$ max\_driver\_mem\_usage $\leq$ 2.0  & \multirow{2}{*}{$\uparrow$} & \multirow{2}{*}{*1.2} & \multirow{2}{*}{Lower:512, Upper:10K}\\
        & and 0.92 $\leq$ avg\_driver\_mem\_usage $\leq$ 2.0 & &  & \\
        \hline
        \multirow{2}{*}{spark.driver.memoryOverhead} & 0.8 $\leq$ max\_driver\_mem\_usage $\leq$ 1.0  & \multirow{2}{*}{$\uparrow$} & \multirow{2}{*}{*1.1} & \multirow{2}{*}{Lower:512, Upper:10K}\\
        & and 0.72 $\leq$ avg\_driver\_mem\_usage $\leq$ 1.0 & &  & \\
        \hline
        \multirow{2}{*}{spark.driver.memoryOverhead} & 0.9 $\leq$ max\_driver\_mem\_usage $\leq$ 1.0  & \multirow{2}{*}{$\uparrow$} & \multirow{2}{*}{*1.1} & \multirow{2}{*}{Lower:512, Upper:10K}\\
        & and 0.6 $\leq$ avg\_driver\_mem\_usage $\leq$ 1.0 & &  & \\

        \bottomrule
    \end{tabular}
    }
    
    \label{tab:rule2}
\end{table*}

We also provide instructions on how to design expert rules for each parameter in the search space. We highlight the following two important parts among all five parts mentioned in Section~\ref{sec:rules}:

a) Condition. To design the condition of an expert rule, we should answer two questions: 1) What is the direct effect of a certain parameter on Spark? and 2) How can we measure this effect using running metrics from Spark? Then, we tune the parameter based on the observed metric values.

We take the parameter \texttt{sparkFilesMaxPartitionBytes} as an example. This parameter directly controls the maximum partition size of input data during mapping, and it influences the number of mappers that process the input data (Question 1). The metric stage\_max\_avg\_input\_run\_time records the runtime of a mapper (Question 2), thus it can be used to check whether the parameter \texttt{sparkFilesMaxPartitionBytes} is properly set. Concretely, a large stage\_max\_avg\_input\_run\_time indicates that the number of mappers is insufficient, so we should increase the number of mappers by reducing \texttt{sparkFilesMaxPartitionBytes}.

b) Step. In Rover, we update the parameter by multiplying it by a constant. The constant value influences the speed to reach the region of near-optimal configurations. i) We apply a large step for parameters where the upper bound is significantly higher than the lower bound. For example, the constant value is 2 or 0.5 for \texttt{sparkFilesMaxPartitionBytes} where the parameter ranges from 16M to 8G. ii) In addition, we suggest using a small step for sensitive parameters. Remind that we use SHAP importance to build the search space. We provide the SHAP importance of 30 parameters above in Table~\ref{tab:shap_value}. For parameters with large SHAP importance (e.g., 9.72 for \texttt{sparkExecutorMemory}), we set the constant value as 1.1 or 0.9 for most cases.

\subsection{Additional Results}
\label{appendix:addition_results}
\subsubsection{Comparison with expert-assisted methods}
As mentioned in Section~\ref{sec:related}, the rule-based methods only provide what parameters to tune rather than how to tune them. 
In this part, we compare Rover with cost model-based methods. We take Ernest~\cite{venkataraman2016ernest} as a representative, which builds a heuristic parametric cost model on some Spark parameters. 
The cost model is trained on previous observations and then suggests the configuration that maximizes the output of the cost model. 
As the total tuning budget of Rover is 50 iterations, we also fit the cost model in Ernest using observations of 50 random configurations.

To ensure a fair comparison, we disable controlled history transfer in Rover. The configuration suggested by Ernest achieves an average memory cost ratio of 21.37\% on 200 internal tasks. As an iterative tuning method, it takes Rover only 5 iterations to achieve that result. Moreover, with the same budget (50 iterations), Rover achieves an average memory cost ratio of 7.15\%, which is 14.22\% lower than Ernest. We attributed this gain to a well-designed set of expert rules and the use of BO algorithm, which models the complex relationship between configurations and performance rather than assuming a simple relationship based on heuristic knowledge.

\subsubsection{Optimization on another target}
To show the generality of Rover, we evaluate Rover using another target (accumulated CPU cost). 
The accumulated CPU cost is computed as the CPU usage multiplied by the runtime, which is also an important metric used in industry.

To support different targets, the users do not need to alter any algorithm parts of Rover. 
Instead, they only need to collect the corresponding performance metric of the configuration, and Rover will automatically model the relationship between those configurations and performance metric. 
The expert rules still work when the target changes from memory cost to CPU cost. 
For history transfer, the users should provide the tuning history on CPU cost instead of memory cost.

\begin{figure}
  \begin{center}
	\includegraphics[width=0.96\linewidth]{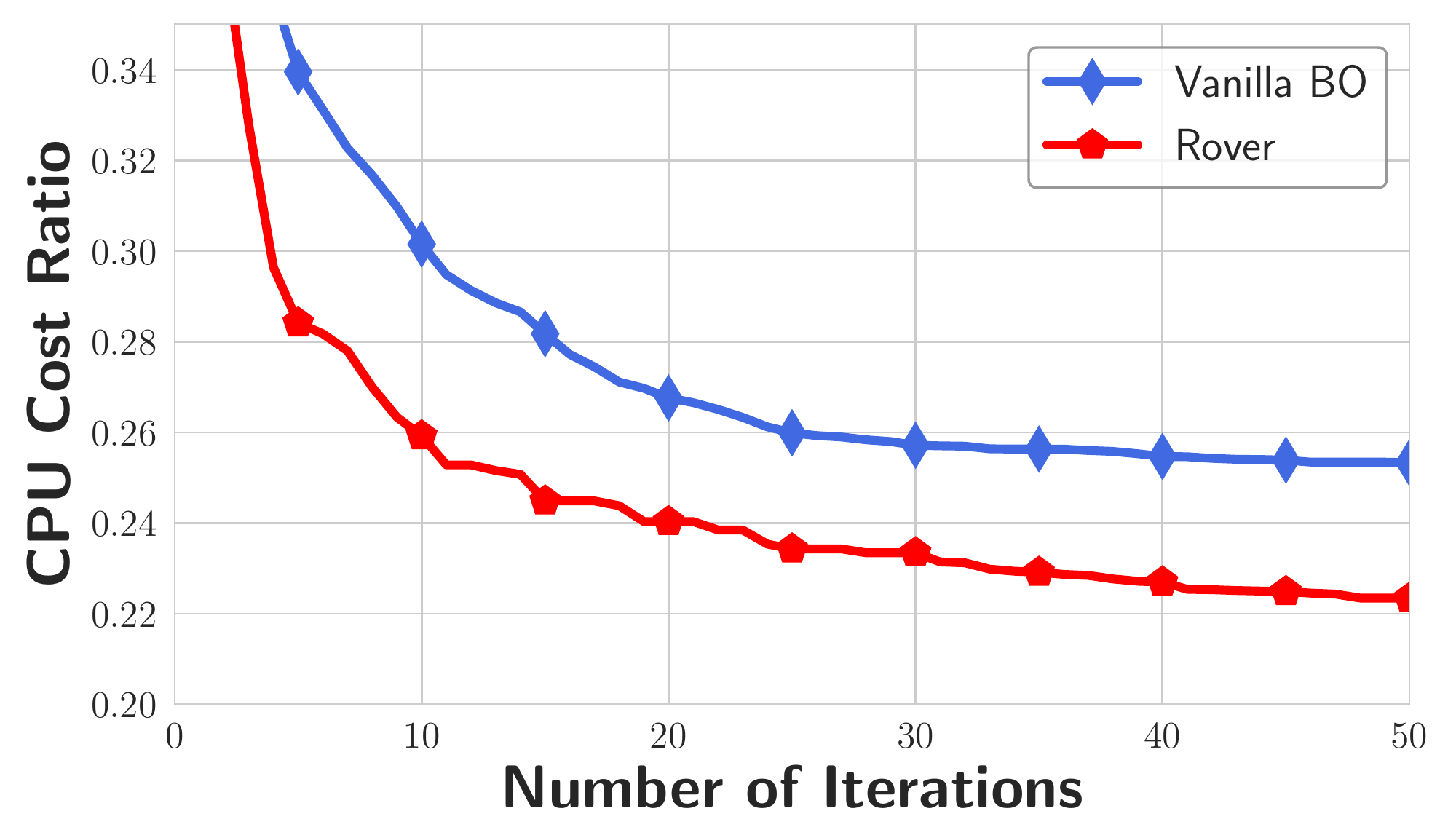}
  \end{center}
  \caption{CPU cost ratio of \sys and vanilla BO.}
 \label{fig:cpu}
\end{figure}

Then, we show the results of Rover using accumulated CPU cost as the target. 
We compare Rover with vanilla BO on 200 internal tasks using the same compact space, and the average results (CPU cost ratio relative to default configurations, the lower the better) are shown in Figure~\ref{fig:cpu}.

We observe that Rover still achieves clear improvement compared with Bayesian optimization (4.22\% and 2.92\% at the 10-th and 50-th iteration), which demonstrates the generality of Rover on different requirements.

\end{document}